\documentclass{article}

     \usepackage[preprint]{neurips_2022}

\usepackage[utf8]{inputenc}     
\usepackage[T1]{fontenc}        
\usepackage{hyperref}           
\usepackage{url}                
\usepackage{booktabs}           
\usepackage{amsfonts}           
\usepackage[super]{nth}         

\usepackage{amsmath}            
\usepackage{amssymb}            
\usepackage[dvipsnames]{xcolor} 
\usepackage{bm}                 
\usepackage{dsfont}             
\usepackage{caption}            
\usepackage{subcaption}         
\usepackage{graphicx}           
\usepackage{microtype}      

\definecolor{auBlue}{RGB}{0,62,92}
\definecolor{auLightBlue}{RGB}{55,160,203}
\definecolor{mydarkblue}{rgb}{0,0.08,0.45}

\hypersetup{ %
    colorlinks=true,
    linkcolor=mydarkblue,
    citecolor=mydarkblue,
    citebordercolor=mydarkblue,
    filecolor=mydarkblue,
    urlcolor=mydarkblue,
    }

\newcommand{\D}{\mathcal{D}}
\newcommand{\E}{\mathbb{E}}

\newcommand{\N}{\mathcal{N}}

\newcommand{\R}{\mathbb{R}}

\newcommand{\dd}{\mathrm{d}}
\newcommand{\norm}[1]{\left\lVert #1 \right\rVert}

\newcommand{\bmu}{\bm{\mu}}

\newcommand{\bsigma}{\bm{\sigma}}
\newcommand{\bSigma}{\mathbf{\Sigma}}

\newcommand\eqdef{\mathrel{\overset{\makebox[0pt]{\mbox{\normalfont\tiny\sffamily def}}}{=}}} 
\newcommand{\trans}{{\intercal}}

\newcommand\bA{\mathbf{A}}
\newcommand\bB{\mathbf{B}}
\newcommand\bC{\mathbf{C}}
\newcommand\bD{\mathbf{D}}

\newcommand\bH{\mathbf{H}}
\newcommand\bI{\mathbf{I}}

\newcommand\bK{\mathbf{K}}

\newcommand\bO{\mathbf{O}}

\newcommand\bV{\mathbf{V}}
\newcommand\bW{\mathbf{W}}

\newcommand\bbf{\mathbf{f}} 

\newcommand\bk{\mathbf{k}}

\newcommand\bbm{\mathbf{m}} 

\newcommand\bo{\mathbf{o}}

\newcommand\bx{\mathbf{x}}

\newcommand\by{\mathbf{y}}

\usepackage[bibliography=common]{apxproof}
\renewcommand{\appendixprelim}{\onecolumn \section{Proofs from main text} In the following we restate all the results of the main paper and provide the associated proofs.} 
\newcommand{\mainbodyrepeatedtheoremend}{\nolinebreak\hfill(Proof in appendix.)}

\theoremstyle{plain}
\newtheoremrep{theorem}{Theorem}[section]
\newtheoremrep{prop}[theorem]{Proposition}
\newtheoremrep{lemma}[theorem]{Lemma}
\newtheoremrep{corollary}[theorem]{Corollary}

\theoremstyle{definition}
\newtheoremrep{definition}[theorem]{Definition}
\newtheoremrep{assumption}[theorem]{Assumption}

\theoremstyle{remark}
\newtheoremrep{remark}[theorem]{Remark}

\newcommand{\codereference}[1]{\small\textbf{Code}: \url{#1}}

\title{Self-Distillation for Gaussian Process Regression and Classification}

\author{%
  Kenneth Borup \\
  Department of Mathematics\\
  Aarhus University\\
  \texttt{kennethborup@math.au.dk} \\
  \And
  Lars Nørvang Andersen \\
  Department of Mathematics\\
  Aarhus University\\
  \texttt{larsa@math.au.dk}
}
\renewcommand{\trans}{{\intercal}}
\usepackage{here}

\DeclareMathOperator{\diag}{diag}
\newcommand{\e}{{\mathrm{e}}}
\DeclareBoldMathCommand\bzero{0}

\begin{document}
\maketitle

\begin{abstract}
We propose two approaches to extend the notion of knowledge distillation to Gaussian Process Regression (GPR) and Gaussian Process Classification (GPC); data-centric and distribution-centric. The data-centric approach resembles most current distillation techniques for machine learning, and refits a model on deterministic predictions from the teacher, while the distribution-centric approach, re-uses the full probabilistic posterior for the next iteration.
By analyzing the properties of these approaches, we show that the data-centric approach for GPR closely relates to known results for self-distillation of kernel ridge regression and that the distribution-centric approach for GPR corresponds to ordinary GPR with a very particular choice of hyperparameters. Furthermore, we demonstrate that the distribution-centric approach for GPC approximately corresponds to data duplication and a particular scaling of the covariance and that the data-centric approach for GPC requires redefining the model from a Binomial likelihood to a continuous Bernoulli likelihood to be well-specified. To the best of our knowledge, our proposed approaches are the first to formulate knowledge distillation specifically for Gaussian Process models.

\smallskip
\codereference{github.com/Kennethborup/gaussian_process_self_distillation}
\end{abstract}

\section{Introduction and Problem Setting}
In this paper, we propose a notion of knowledge distillation (KD) in a Gaussian process (GP) context. Despite numerous research and applications of (extensions of) KD in the context of deep learning, research on knowledge distillation for other methods of machine learning than deep learning is lacking. Recent work by \citet{Borup2021, mobahi2020selfdistillation, phuong2019towards, Frosst2018DistillingTree} amongst others, provide results on KD for simplified settings, such as for (kernel ridge) regression. A natural extension to these results is to investigate KD for GPs, but since no prior work has been performed in this context, we initially propose multiple approaches to how one could define knowledge distillation for GPs. In this paper, we restrict our analysis to self-distillation, i.e. where the student and teacher models are of an identical class, which in practice corresponds to the kernel function being identical for both models.

Unlike e.g. deep learning, where samples and models are deterministic during training, GPs include stochasticity in the form of a prior and thus posterior predictive distribution. In ordinary GP regression, one defines a prior, and based on a set of observations $\D$, a posterior model is obtained which can be used for inference on previously unseen samples. We propose two different and straightforward approaches to self-distillation of GPs: \emph{data-centric} and \emph{distribution-centric} (see Figure \ref{fig:methods}).

\begin{figure*}[htbp]
    \centering
    \includegraphics[width=0.9\linewidth]{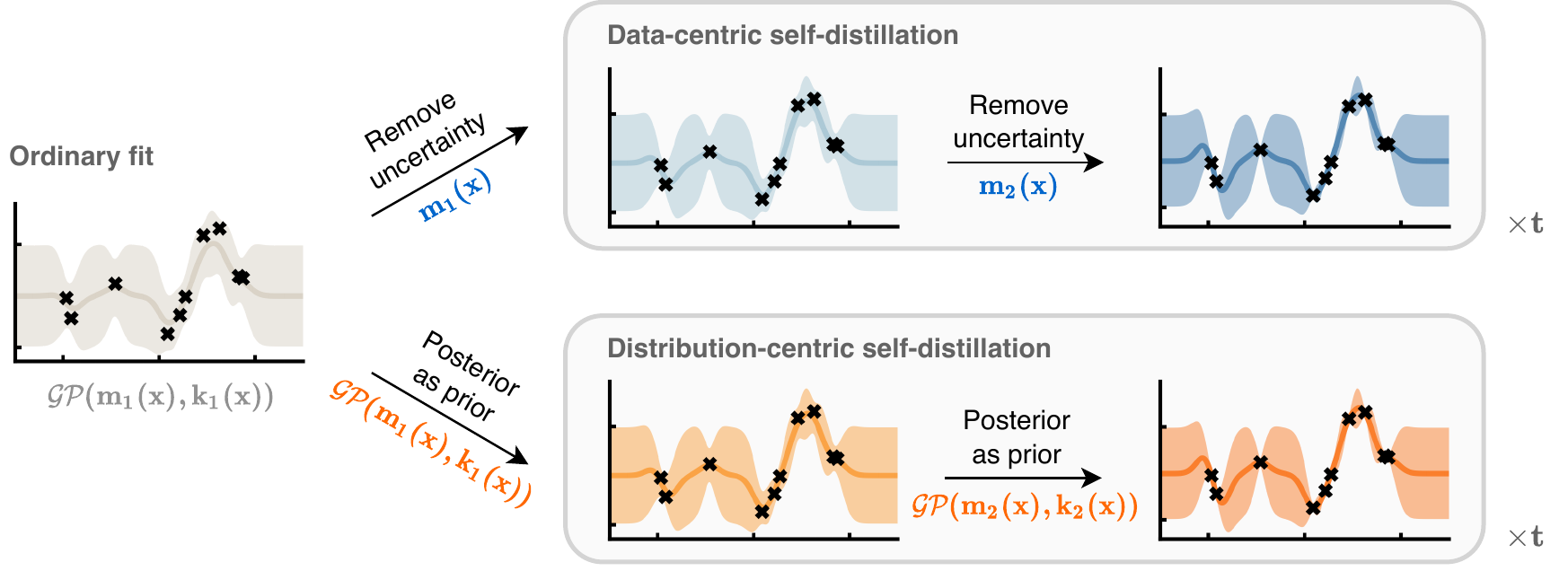}
    \caption{Conceptual illustration of our two proposed methods of performing distillation for Gaussian processes. In data-centric self-distillation, we re-use the mean function of the previous solution in the next step, and in distribution-centric self-distillation, we re-use the full distribution.}
\end{figure*}

\paragraph{Data-centric self-distillation.}
In data-centric distillation, we treat the output predictions of a model at step $t$ as the input to the model at step $t+1$. This procedure aligns well with most current distillation methods known from deep learning \citep{hinton2015distilling}. However, unlike in deep learning, the output of our models is not merely scalars, but distributions. We propose to discard the stochasticity of our predictions at step $t$ and merely keep the mean predictions for training of the model at step $t+1$. We show that for GP regression the mean after self-distillation is closely related to distillation with kernel ridge regression as in \citet{Borup2021, mobahi2020selfdistillation}. Furthermore, we argue that for distillation in a classification setting, one needs to adapt the likelihood function to support continuous targets rather than binary targets. We address this by utilization of the continuous Bernoulli distribution \citep{Loaiza-Ganem2019TheAutoencoders}. We investigate data-centric distillation for regression in Section \ref{sec:data_sd} and for classification in Section \ref{sec:data_gpc}.

\paragraph{Distribution-centric self-distillation.}
In distribution-centric distillation, we replace the prior at a given step, $t$, with the posterior of the previous step, $t-1$. In GP regression the posterior distribution is a GP itself, and we can directly consider it as a prior for the succeeding step. This yields an iteratively refined and data-dependent prior with each distillation step. For classification, more care has to be taken as the posterior is not known analytically. However, by utilizing the Laplace approximation of the posterior as the prior of the latent GP model, we are able to define a meaningful notion of distribution-centric distillation in the classification setup as well. We investigate distribution-centric for regression in Section \ref{sec:distribution_sd} and for classification in Section \ref{sec:distribution_gpc}.

\paragraph{Combined and extended directions.}
As is commonly practiced in the literature, one can consider the extension of the data-centric distillation approach, where we not only consider the posterior predictions of the previous step but rather consider a convex combination (weighted by $\alpha \in (0,1)$) of these posterior predictions and the original observations.
Furthermore, since the data-centric and distribution-centric approaches are not mutually exclusive, we could consider a combined method, with an inner and outer distillation loop. In the outer loop the prior is replaced by the posterior of the preceding step as in distribution-centric distillation, but in the inner loop, the posterior is distilled a set number of steps using data-centric distillation, with the prior fixed in the outer loop.
We leave such intricate, but intriguing setups for future research.

We summarize our \textbf{contributions} as:
\begin{itemize}
    \item We propose two different approaches to self-distillation for both GP regression and GP classification, reusing either the mean predictions (data-centric) or full predictive distribution (distribution-centric) of the previous iteration of the model.
    \item We show a connection between data-centric self-distillation for GP regression and self-distillation of kernel ridge regression. Furthermore, we prove that distribution-centric self-distillation for GP regression is equivalent to ordinary GP regression with a particular choice of hyperparameter.
    \item We find that a naïve approach to data-centric self-distillation for GP classification yields a misspecified model due to the continuous form of the predictions, and we propose to alleviate this by utilization of the continuous Bernoulli distribution. Furthermore, we find that distribution-centric self-distillation can be efficiently approximated by a particular scaling of the covariance function for any number of distillation steps.
\end{itemize}

The structure of the paper is as follows. In Section \ref{sec:preliminaries} we present preliminary results on GP regression and GP classification. Next, we focus on self-distillation for GP regression in Section \ref{sec:gpr_sd}, and in particular on data-centric distillation in Section \ref{sec:data_sd} and distribution-centric distillation in Section \ref{sec:distribution_sd}. Afterward, in Section \ref{sec:gpc_sd} we investigate the classification setup and in particular the data-centric setting in Section \ref{sec:data_gpc} and distribution-centric setting in Section \ref{sec:distribution_gpc}. Finally, we relate our work to existing research in Section \ref{sec:related_works}. Additionally, in the appendix we provide additional mathematical details, implementation details, experiment ablations, as well as all proofs.

\begin{figure*}[t]
    \centering
    \hspace{0.75cm}
    \begin{subfigure}[b]{0.22\textwidth}
        \centering
        \includegraphics[width=\textwidth]{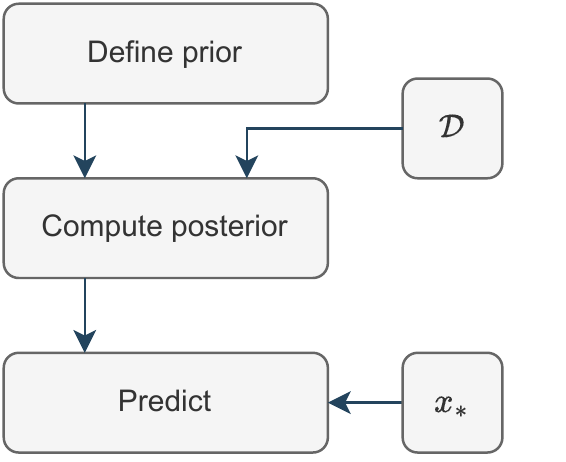}
        \caption{Classical.}
        \label{fig:gp_classical}
    \end{subfigure}
    \hfill
    \begin{subfigure}[b]{0.34\textwidth}
        \centering
        \includegraphics[width=\textwidth]{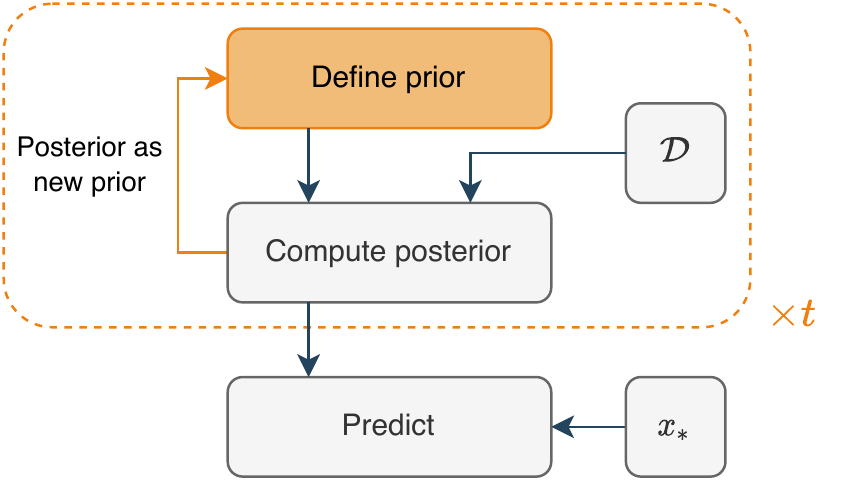}
        \caption{Distribution-centric.}
        \label{fig:gp_distribution}
    \end{subfigure}
    \hfill
    \begin{subfigure}[b]{0.34\textwidth}
        \centering
        \includegraphics[width=\textwidth]{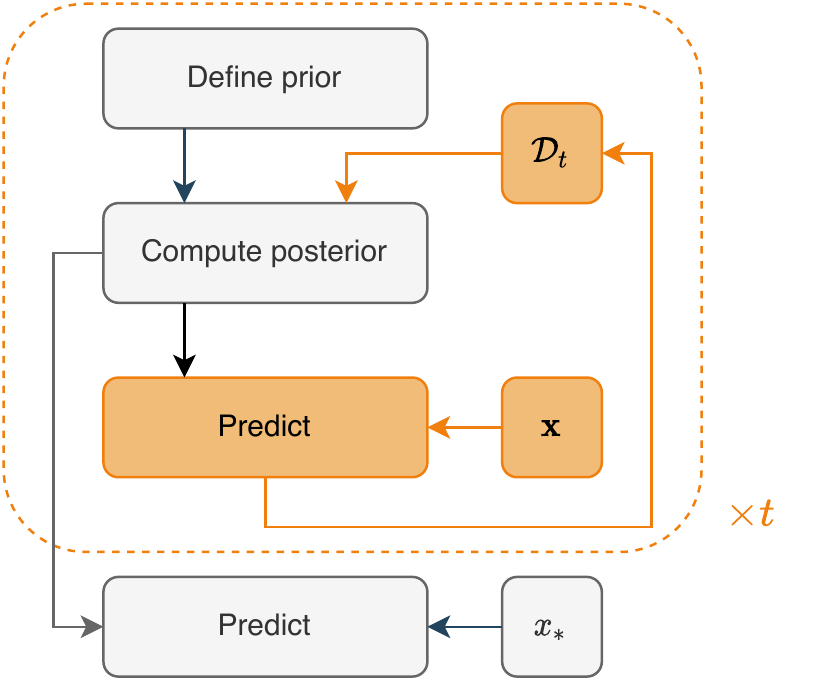}
        \caption{Data-centric.}
        \label{fig:gp_data}
    \end{subfigure}
    \caption{In (\subref{fig:gp_classical}) we illustrate a typical procedure to fit a GP to an observed dataset, $\D = \{(x_i, y_i)\}_{i=0}^n$, and compute the output distribution for some unseen sample $x_\ast$. In (\subref{fig:gp_distribution}) we illustrate the distribution-centric self-distillation procedure, where we reuse the posterior as a prior. In (\subref{fig:gp_data}), we illustrate the data-centric self-distillation procedure, where we iteratively replace the training observations $\D_t = \{(x_i, y_{t,i})\}_{i=0}^n$ with the posterior predictions of the current model, $\D_{t+1} = \{(x_i, y_{t+1,i})\}_{i=0}^n$, and refit the model to these samples.}
    \label{fig:methods}
\end{figure*}
\section{Preliminaries}\label{sec:preliminaries}
Our starting point is the standard set-up for Gaussian Processes regression. These results are well-known (see e.g. \citet{Rasmussen2006}).

\subsection{Gaussian Process regression}\label{sec:GPR}
We assume a generic input-output pair $(\bx,y) \in \R^d \times \R$ is related by $y = f(\bx) + \varepsilon$, where $f(\cdot)$ is a function, which we wish to infer, based on the assumption that $f$ is a random function whose prior distribution is $f \sim \mathcal{GP}(m, k)$ for mean function $m: \R^d \to \R$ and positive semi-definite covariance function (often called kernel) $k: \R^d \times \R^d \to \R$, and $\varepsilon$ is an independent noise term with $\varepsilon \sim \N(0,\gamma)$. For notational convenience we will assume that our input is univariate, i.e. $d=1$. For a training set 
$\mathcal{D}  = \{ (x_i, y_i) \mid i = 1,\dots,N\}$ and test set $(x_{\ast 1},\dots,x_{\ast M})^\trans$, the first task is to describe the posterior distribution of $\bbf_\ast = (f(x_{\ast 1}),\dots,f(x_{\ast M}))^\trans$ given $\mathcal{D}$. 
The posterior predictive distribution of $\bbf_\ast$ given $\mathcal{D}$ is 
\begin{align*}
    \bbf_\ast &\mid \bx, \by, \bx_\ast \sim \N(\bmu_\ast, \bSigma_\ast), \quad \text{where} \\
    \bmu_\ast &= m(\bx_\ast) + k(\bx_\ast, \bx^\trans)(\bK + \gamma \bI_N)^{-1}\left(\by - m(\bx) \right), \\
    \bSigma_\ast &= k(\bx_\ast, \bx_\ast^\trans) - k(\bx_\ast, \bx^\trans)(\bK + \gamma \bI_N)^{-1} k(\bx,\bx_\ast^\trans).
\end{align*}
and where $\bx$ and $\bx_\ast$ are stacked training and test input, respectively.
In the expressions for $\bmu_\ast$ and $\bSigma_\ast$, we have used the notation 
\begin{align*}
    g(\bx,\by^\trans) &= \left\{ g(x_i,y_j) \right\}_{i,j=1}^{N,M} \in \mathbb{R}^{N \times M}
\end{align*}
where  $g: \mathbb{R}^2 \ni (x,y) \mapsto g(x,y) \in \mathbb{R}$, and $\bx = (x_1,\dots,x_N)^\trans \in \mathbb{R}^N$, $\by = (y_1,\dots,y_M)^\trans\in \mathbb{R}^M$ are  column vectors. Notice, that if either argument is a scalar we have 
\begin{align*}
    g(x,\bx^\trans) &= \left( g(x,x_1),\dots, g(x,x_N) \right) \in \mathbb{R}^{1 \times N} \\
    g(\bx,y) &= \left( g(x_1,y),\dots, g(x_N,y) \right)^\trans \in \mathbb{R}^{N \times 1}
\end{align*}
where $g(x,\bx^\trans)$ is a row-vector and $g(\bx,y)$ a column vector.

Throughout the paper, we will write $\bK$ for the matrix $\bK = k(\bx, \bx^\trans)$ which we assume is invertible. In practice, if $\bK$ is not invertible, one can add a small constant to the diagonal. Furthermore, we will typically omit the conditioning on $\bx$ and $\bx_\ast$ for ease of exposition.

\subsection{Gaussian Process classification}\label{sec:gpc_prelim}
We follow the usual setup for classification with Gaussian Processes in which each input-output pair $(\bx,y)$ is related by the assumption that the conditional distribution $y \mid f(\bx)$ is a Bernoulli distribution with probability $\sigma(f(\bx))$ where $\sigma(\cdot)$ is the logistic function and the prior distribution of $f$ is $\mathcal{GP}(m,k)$. 
For a training set $\mathcal{D} = \{ (x_i, y_i) \mid i = 1,\dots,N\}$ we denote $\bbf = (f_n)_{n=1}^N = (f(\bx_1),\dots,f(\bx_N))^\trans$ and $\by = (y_1, \dots, y_n)\trans$ so that 
\begin{align}
    \by \mid \bbf \sim \prod_{n=1}^N \sigma(f_n)^{y_n}(1-\sigma(y_n))^{1-y_n} \, .
    \label{eq:GPdef}
\end{align}
The log posterior has the form
\begin{align*}
    \psi(\bbf) \eqdef \log p(\bbf \mid \by) \underset{\bbf}{\propto} \log p(\bbf) + \log p(\by \mid \bbf),
\end{align*}
where $p(\bbf)$  is the pdf of the multivariate Gaussian distribution and it follows from \eqref{eq:GPdef} that
\begin{align*}
    \log p(\by \mid \bbf) = \by^\trans \bbf - \sum_{n=1}^N \log (1+\exp(f_n)) \ .
\end{align*}
 Unlike the classification case, the posterior distribution $\bbf \mid \by$ is not analytically tractable and needs to be approximated. 
This issue is not specific to distillation but arises in any practical application involving Gaussian process classification, and a number of approaches have been suggested for obtaining an approximation of the posterior distribution. In this paper, we will focus on the Laplace approximation, where the posterior distribution is approximated by
\begin{align*}
    q(\bbf \mid \by) = \mathcal{N}\left( \bbf\mid  \hat{\bbf} , (\bK^{-1} + \bW)^{-1} \right)
\end{align*}
where $\hat{\bbf} = (\hat{f}_n)_{n=1}^N$ is the mode of $\psi(\bbf)$ and 
$\bW$ is the $N \times N$ matrix $\diag_n \{ \sigma(\hat{f}_n) (1 - \sigma(\hat{f}_n)) \}$.
The approximate posterior predictive distribution is found in \citet[See Section 3.4.2]{Rasmussen2006} in the case where $m = 0$ and for a general prior mean $m$, we have $p(f_\ast \mid \bx, \by) \approx \mathcal{N}(\bmu_\ast,\bSigma_\ast)$ where
\begin{align}
    \bmu_\ast &= \E_q[\bbf_\ast \mid \by, \bx_\ast] \notag \\
    &= m(\bx_\ast) + k(\bx_\ast, \bx^\trans) \bK^{-1} (\hat{\bbf} - m(\bx)) \label{eq:gppredmean} \\
    \bSigma_\ast &= \mathrm{Cov}_q(\bbf_\ast \mid \by, \bx_\ast) \notag \\
    \quad &= k(\bx_\ast, \bx_\ast^\trans) - k(\bx_\ast, \bx^\trans) (\bK + \bW^{-1})^{-1}k(\bx, \bx_\ast^\trans). \label{eq:gppredbSigma}
\end{align}
See Section \ref{sec:GPC} for additional details on Gaussian Process classification. We return to the classification setting in Section \ref{sec:gpc_sd}. 

\section{Self-distillation for GP Regression}\label{sec:gpr_sd}
In the following we consider data- and distribution-centric self-distillation for GP regression (GPR) separately.

\subsection{Data-centric Self-Distillation}\label{sec:data_sd}

\begin{figure*}[htbp]
    \centering
    \begin{subfigure}[b]{0.49\textwidth}
        \centering
        \includegraphics[width=1\linewidth]{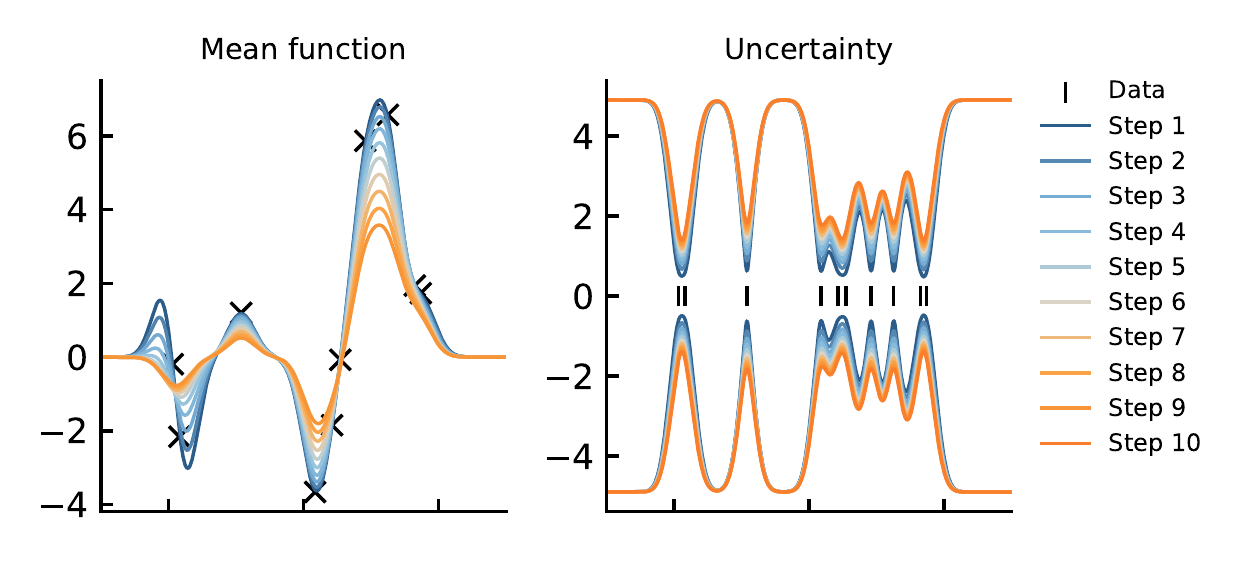}
        \caption{Data-centric distillation for ten steps.}
        \label{fig:gpr_data_example}
    \end{subfigure}
    \hfill
    \begin{subfigure}[b]{0.49\textwidth}
        \centering
        \includegraphics[width=1\linewidth]{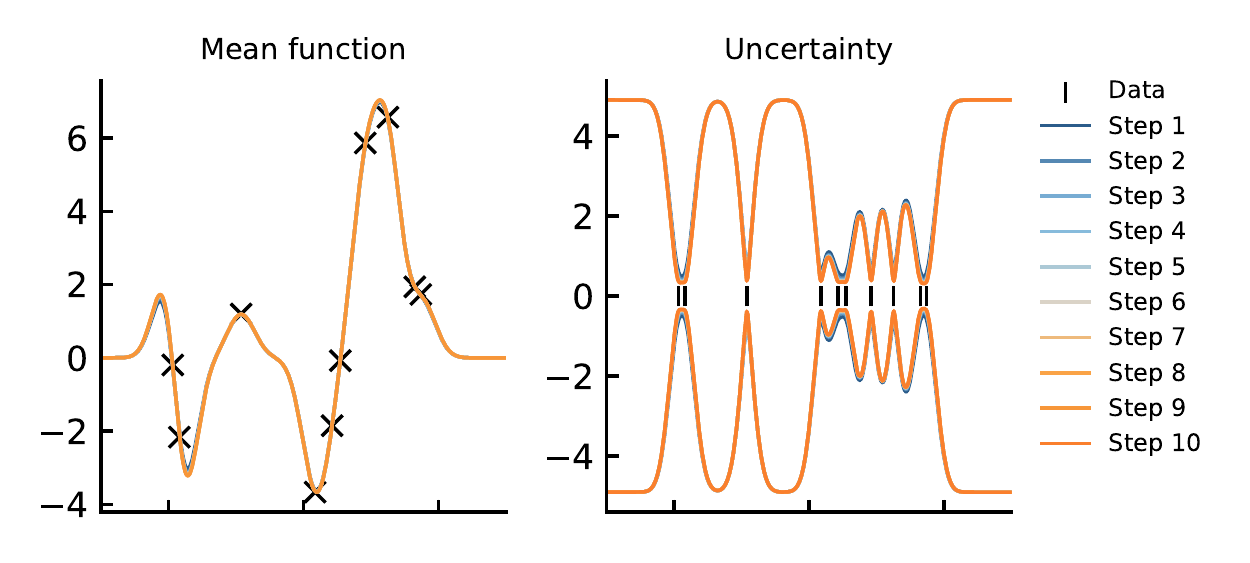}
        \caption{Distribution-centric distillation for ten steps.}
        \label{fig:gpr_distribution_example}
    \end{subfigure}
    \caption{Ten steps of self-distillation with fixed hyperparameters and $(\gamma_1, \dots, \gamma_{10}) = (0.1, \dots, 1)$ for both the data-centric and distribution-centric approach. We plot both the mean function and the 2.5 and 97.5 percentiles of the uncertainty estimate at each step.}
\end{figure*}

For the general setup of data-centric distillation, we consider a mean-zero prior distribution, $f \sim \mathcal{GP}(0,k)$,
and the posterior distribution on $\bx = (x_1, \dots, x_N)^\trans$ at step $t \geq 1$ as $\bbf_t \mid \by_{t-1} \sim \N(\bmu_t, \bSigma_t)$, where
\begin{align*}
    \bmu_t &= \bK(\bK + \gamma_t \bI_N)^{-1}\by_{t-1}, \\
    \bSigma_t &= \bK - \bK(\bK + \gamma_t \bI_N)^{-1}\bK,
\end{align*}
and $\gamma_t > 0$ for $t \geq 1$.
Data-centric self-distillation corresponds to using the mean predictions $\bmu_t$ in place of $\by_t$ in fitting of the succeeding step, thereby fitting the subsequent model to the mean predictions of the current model.
In this setting the behavior of the mean function becomes equivalent to that of \citet{mobahi2020selfdistillation} and \citet{Borup2021} as evident from the following result.
\begin{proprep}\label{prop:scalar_data}
    Define $\by_t \eqdef \bmu_t$ and $\by_0 \eqdef \by$.
    Then it holds for any $t \geq 1$ that $\bbf_t \mid \bmu_{t-1} \sim \N\left(\bmu_t, \bSigma_t \right)$, where
    \begin{align} \label{eq:kernel_sd}
        \by_t &= \left(\prod_{s=1}^{t}\bK(\bK + \gamma_s \bI_N)^{-1}\right)\by, \\
        \E[\bbf_{t,\ast} \mid \bmu_{t-1}] &= k(\bx_\ast, \bx)(\bK + \gamma_t \bI_N)^{-1} \by_{t-1}, \label{eq:kernel_sd_pred}
    \end{align}
    and $\bSigma_t = \bK - \bK(\bK + \gamma_t \bI_N)^{-1}\bK$ as well as $\by_t$ is only affected by the choice of $\gamma_t$.
\end{proprep}
\begin{proof}
    We assume the model $y_{t-1} = f_t(\bx) + \varepsilon_t$, with $f_t \sim \mathcal{GP}(0,k)$, and $\varepsilon_t \sim \N(0, \gamma_t)$ for all $t \geq 1$. This yields
    \begin{align*}
        p(\bbf_t) &= \N(\bbf_t \mid 0, \bK) \\
        p(\by_{t-1} \mid \bbf_t) &= \N(\by_{t-1} \mid \bbf_t, \gamma_t\bI_N) \\
        p(\by_{t-1}) &= \int p(\by_{t-1} \mid \bbf_t)p(\bbf_t) \,\mathrm{d}\bbf_t = \N(\by_{t-1} \mid 0, \bK + \gamma_t\bI_N),
    \end{align*}
    where we use the usual bolded notation for functions evaluated in $\bx_1, \dots, \bx_N$; i.e. $\by_t = (y_t(\bx_1),\dots,y_t(\bx_N))^\trans \in \R^N$ and $\bbf_t = (f_t(\bx_1),\dots,f_t(\bx_N))^\trans \in \R^N$. Which combined yields the prior
    \begin{align*}
        \begin{bmatrix}
            \by_{t-1} \\
            \bbf_{t}
        \end{bmatrix}
        \;\sim\; \N\bigg(
        \begin{bmatrix}
            \mathbf{0} \\
            \mathbf{0}
        \end{bmatrix},
        \begin{bmatrix}
            \bK + \gamma_t\bI    & \bK  \\
            \bK                  & \bK
        \end{bmatrix}
        \bigg).
    \end{align*}
    Thus, following the standard results for conditional multivariate Gaussian distributions, we get the posterior
    \begin{align}\label{eq:posterior_eq_kernel_proof}
        \bbf_{n} \mid \by_{t-1} &\sim \N\Big(\bK(\bK + \gamma_t\bI_N)^{-1}\by_{t-1},\; \bK - \bK(\bK + \gamma_t\bI_N)^{-1}\bK \Big).
    \end{align}
    Recall that we use $\by_t \eqdef \bmu_t = \E[\bbf_{n} \mid \by_{t-1}]$, and by iteratively inserting $\by_{t-1} = \bK(\bK + \gamma_{t-1}\bI_N)^{-1}\by_{t-2}$ into \eqref{eq:posterior_eq_kernel_proof}, we get
    \begin{align*}
        \by_t = \E[\bbf_t \mid \by_{t-1}] &= \bK(\bK + \gamma_t\bI_N)^{-1}\by_{t-1} \\
        &= \bK(\bK + \gamma_t\bI_N)^{-1}\bK(\bK + \gamma_{t-1}\bI_N)^{-1}\by_{t-2} \\
        &= \bA_t \bA_{t-1}\by_{t-2} \\
        &= \left(\prod_{m=1}^n \bA_m\right) \by_0,
    \end{align*}
    where we use the notation $\bA_i = \bK(\bK + \gamma_i\bI_N)^{-1}$. Note, the claim for $\E[f_t(x)]$ follows directly from analogous results for arbitrary $\bx$.
\end{proof}

It can be shown that the behavior of $\by_t$ as $t$ increases corresponds to a progressively sparsifying the underlying basis functions by shrinking the basis functions with smallest eigenvalues the most, thereby imposing a regularization effect on the predictions \citep{mobahi2020selfdistillation, Borup2021}.
Furthermore, since $\by_t$ only depend on $t$ through $\{\gamma_s\}_{s=1}^t$, once can efficiently compute \eqref{eq:kernel_sd} for any choice of $t$, by using a spectral decomposition of $\bK$.

Furthermore, following \citet{Borup2021} the results can be extended to include a weighting (by $\alpha$) of the original training observations and (by $1-\alpha$) of the predictions from the previous iteration of the model. 

\subsection{Distribution-centric Self-Distillation}\label{sec:distribution_sd}
The central observation of the distribution-centric approach is that 
the posterior distribution is itself a Gaussian Process, namely $\mathcal{GP}(m_\ast, k_\ast)$, where
\begin{align*}
    m_\ast(x) &= m(x) + k(x, \bx^\trans)\left(\bK + \gamma \bI_N\right)^{-1}\left(\by - m(\bx) \right) \\
    k_\ast(x,y) &= k(x, y) - k(x, \bx^\trans)\left(\bK + \gamma \bI_N\right)^{-1} k(\bx,y) \, .
\end{align*}
Since the posterior distribution is a Gaussian process, we may ''distill'' the posterior distribution, by using it as a prior distribution in our initial set-up, and we may iterate this procedure for any finite number of steps, see Figure 1 (\subref{fig:gp_distribution}). Furthermore, we may select the noise terms $\gamma_t$ to be different in each step,
and we see that the distillation procedure leads to a \textit{series} of Gaussian Processes $\mathcal{GP}(m_t,k_t)$ where the mean- and covariance function satisfy the recursions 
\begin{align}
m_{t+1}(x)  &= m_t(x) + \label{eq:recmean} k_t(x,\bx^\trans) [\bK_t + \gamma_t \bI_N]^{-1} ( \by - m_t(\bx) ) \\
k_{t+1}(x,y) &= k_t(x,y) -  \label{eq:recvar} k_t(x,\bx^\trans) [\bK_t + \gamma_t \bI_N]^{-1} k_t(\bx,y) 
\end{align}
where $\bK_t \eqdef k_t(\bx,\bx^\trans)$,
$m_0(x) \eqdef 0$, $k_0(x,y) \eqdef k(x,y)$ and $\gamma_t > 0$, $t=0,1,\dots$.

The solutions to the above recursions \eqref{eq:recmean} and \eqref{eq:recvar} are given 
in Theorem \ref{th:mainrec} below, where we show that iterating the recursions $t$ times with
penalty parameters $\gamma_0,\gamma_1,\dots,\gamma_{t-1}$ yields the same result as performing a single iteration with penalty parameter $1/\gamma_{t-1}^-$ where $\gamma_{t}^- \eqdef \sum_{s=0}^{t} 1 / \gamma_s$ and $\gamma_{-1}^{-} \eqdef 0$.
To show this, we first need a lemma.

\begin{lemmarep}
    Let
    $\lambda_i^{(t)}$, for $i = 1,\dots,N$, denote the eigenvalues of $\bK_t$. Write
    $\bK_0 = \bO \diag_i ( \lambda_i^{(0)} ) \bO^\trans$ for the spectral decomposition of $\bK_0$. Then we have for $t = 0,1,\dots$ that
    \begin{align}
        \bK_{t} = \bO \underset{i}{\diag} \left( \lambda_i^{(0)}\zeta_i^{(t)} \right) \bO^\trans \label{eq:recbK}
    \end{align}
    where $\zeta_i^{(t)} \eqdef \frac{1}{\lambda_i^{(0)} \gamma_{t-1}^- + 1}$. For $k_t(x,\bx^\trans)$ and $m_t(\bx)$ we have
    \begin{align}
        k_t(x,\bx^\trans) &= k_0(x,\bx^\trans) \bO \underset{i}{\diag} \left(\zeta_i^{(t)}\right)\bO^\trans \label{eq:reckn} \\
        m_t(\bx) &= \bO \underset{i}{\diag} \left(\lambda_i^{(0)}\gamma_{t-1}^- \zeta_i^{(t)} \right)\bO^\trans \by. \label{eq:recmn}
    \end{align}
\end{lemmarep}

\begin{proof}
    First we observe that \eqref{eq:recvar} gives the following recursion for $\bK_t$: 
    \begin{align}\label{eq:recbK_2}
        \bK_{t+1} = \bK_t - \bK_t ( \bK_t + \gamma_t \bI_N)^{-1} \bK_t
    \end{align}
    If we let $(\lambda,\bo)$ be an eigenvalue-eigenvector pair for $\bK_t$, it follows from \eqref{eq:recbK_2}, that 
    $(\lambda - \lambda^2/(\lambda + \gamma_t),\bo)$ is an eigenvalue-eigenvector pair for $\bK_{t+1}$, so that we have the following recursion for eigenvalues of the $\bK_t$-matrices:
    $$
    \lambda_{i}^{(t+1)} = \lambda_i^{(t)} - \frac{\left(\lambda_i^{(t)}\right)^2}{\lambda_i^{(t)} + \gamma_t} = \frac{\lambda_i^{(t)} \gamma_t}{\lambda_i^{(t)} + \gamma_t} \qquad i = 1,\dots,N \, .
    $$
    From this it can be easily verified by induction, that 
    $$
    \lambda_{i}^{(t)} = \frac{\lambda_i^{(0)}}{\lambda_i^{(0)} \gamma_{t-1}^- + 1} \qquad i = 1,\dots,N \, ,
    $$
    and we see that the solution to the recursion \eqref{eq:recbK} is
    \begin{align}
        \bK_{t} = \bO \underset{i}{\diag} \left( \frac{\lambda_i^{(0)} }{ \lambda_i^{(0)} \gamma_{t-1}^- + 1} \right) \bO^\trans \label{eq:bKrekusion} \, .
    \end{align}
    Notice that \eqref{eq:bKrekusion} trivially holds for $t=0$. Next, we note the following consequences of \eqref{eq:bKrekusion}:
    Firstly, it follows from \eqref{eq:recbK_2} that
    \begin{align}
        (\bK_t + \gamma_t \bI)^{-1} = \bO \underset{i}{\diag} \left( \frac{1}{ \frac{\lambda_i^{(0)} }{ \lambda_i^{(0)} \gamma_{t-1}^- + 1} + \gamma_t} \right) \bO^\trans = \bO \underset{i}{\diag} \left( \frac{\lambda_i^{(0)} \gamma_{t-1}^- +1}{ \gamma_t ( \lambda_i^{(0)} \gamma_t^- +1)} \right) \bO^\trans \, . \label{eq:bKinvbK}
    \end{align}
    Secondly, combining \eqref{eq:bKrekusion} and \eqref{eq:bKinvbK} gives
    \begin{align}
        \bK_t (\bK_t + \gamma_t \bI)^{-1}  = (\bK_t + \gamma_t \bI)^{-1} \bK_t = 
        \bO \underset{i}{\diag} \left( \frac{ \lambda_i^{(0)} }{ \gamma_t ( \lambda_i^{(0)} \gamma_t^- +1)} \right) \bO^\trans \, .
        \label{eq:bKbKinvbK}
    \end{align}
    Finally, we see that
    \begin{equation}
        \begin{split}
            \bI_N - [\bK_t + \gamma_t \bI_N]^{-1}\bK_t &=
            \bO \underset{i}{\diag} \left( 1 -  \frac{\lambda_i^{(0)} /\gamma_t }{ \lambda_i^{(0)} \gamma_{t}^- + 1} \right)\bO^\trans \\
            &=
            \bO \underset{i}{\diag} \left( \frac{\lambda_i^{(0)} \gamma_{t-1}^- + 1}{ \lambda_i^{(0)} \gamma_{t}^- + 1} \right)\bO^\trans     \, .
        \end{split}
        \label{eq:bIminusbKinv}
    \end{equation}
    Now, we can turn our attention to $k_t(x,\bx^\trans)$. Here we have
    \begin{align*}
        k_{t+1}(x,\bx^\trans) = & \big(k_{t+1}(x,x_1),k_{t+1}(x,x_2),\dots,k_{t+1}(x,x_N)\big)  \\
        = & \big(k_t(x,x_1) - k_t(x,\bx^\trans) [\bK_t + \gamma_t \bI_N]^{-1} k_t(\bx,x_1),\dots ,\\
        & k_t(x,x_N) - k_t(x,\bx^\trans) [\bK_t + \gamma_t \bI_N]^{-1} k_t(\bx,x_N) \big) \\
        = & k_t(x,\bx^\trans) -  \big(k_t(x,\bx^\trans) [\bK_t + \gamma_t \bI_N]^{-1} k_t(\bx,x_1),\dots, \\
        & k_t(x,\bx^\trans) [\bK_t + \gamma_t \bI_N]^{-1} k_t(\bx,x_N)\big) \\
        = & k_t(x,\bx^\trans) - k_t(x,\bx^\trans) [\bK_t + \gamma_t \bI_N]^{-1} \bK_t \\
        = & k_t(x,\bx^\trans) \left(\bI_N - [\bK_t + \gamma_t \bI_N]^{-1}\bK_t \right)
    \end{align*}
    Iterating the recursion for $k_t(x,\bx)$ above and using \eqref{eq:bIminusbKinv} we obtain for $t=1,\dots,$
    \begin{equation}
        \begin{split}
            k_{t}(x,\bx^\trans) &= k_0(x,\bx^\trans) \prod_{j=0}^{t-1} \bO \underset{i}{\diag} \left( \frac{\lambda_i^{(0)} \gamma_{j-1}^- + 1}{ \lambda_i^{(0)} \gamma_{j}^- + 1} \right)\bO^\trans \\
            &= k_0(x,\bx^\trans) \bO \underset{i}{\diag} \left( \prod_{j=0}^{t-1} \frac{\lambda_i^{(0)} \gamma_{j-1}^- + 1}{ \lambda_i^{(0)} \gamma_{j}^- + 1} \right)\bO^\trans \\
            &= k_0(x,\bx^\trans) \bO \underset{i}{\diag} \left(  \frac{ 1}{ \lambda_i^{(0)} \gamma_{t-1}^- + 1} \right)\bO^\trans \, .
        \end{split}
    \label{eq:recfork}
    \end{equation}
    Next, we prove \eqref{eq:recmn}. First, we note that \eqref{eq:recmean} gives the following recursion for $m_{t+1}(\bx)$:
    \begin{equation}
        \begin{split}
            m_{t+1}(\bx) &=  m_t(\bx) + k_t(\bx,\bx^\trans) [k_t(\bx,\bx^\trans) + \gamma_t \bI_N]^{-1} ( \by - m_t(\bx) ) \\
            &=  m_t(\bx) + \bK_t [\bK_t + \gamma_t \bI_N]^{-1} ( \by - m_t(\bx) ) \\
            &=  (\bI_t - \bK_t [\bK_t + \gamma_t \bI_N]^{-1}) m_t(\bx) + \bK_t [\bK_t + \gamma_t \bI_N]^{-1}\by
            \end{split}
        \label{eq:mnrec}  
    \end{equation}
    It follows by induction that
    \begin{align}
        m_t(\bx) = \bO \underset{i}{\diag} \left( \frac{\gamma^-_{t-1} \lambda_i^{(0)} }{\gamma^-_{t-1} \lambda_i^{(0)} + 1} \right) \bO^\trans \by
    \label{eq:mninduct}
    \end{align}
    The claim is trivial for $t=0$. Assuming \eqref{eq:mninduct} is true and using 
    \eqref{eq:mnrec} together with \eqref{eq:bKinvbK} and \eqref{eq:bIminusbKinv}, we find
    \begin{align*}
        & m_{t+1}(\bx)\\
        &= \bO \underset{i}{\diag} \left( \frac{\lambda_i^{(0)} \gamma_{t-1}^- + 1}{ \lambda_i^{(0)} \gamma_{t}^- + 1} \right)\bO^\trans \bO \underset{i}{\diag} \left( \frac{\gamma^-_{t-1} \lambda_i^{(0)} }{\gamma^-_{t-1} \lambda_i^{(0)} + 1} \right) \bO^\trans \by
        + \bO \underset{i}{\diag} \left( \frac{ \lambda_i^{(0)} }{ \gamma_t ( \lambda_i^{(0)} \gamma_t^- +1)} \right) \bO^\trans \by \\
        &= \bO \underset{i}{\diag} \left( \frac{ \gamma_{t-1}^- \lambda_i^{(0)}}{ \lambda_i^{(0)} \gamma_{t}^- + 1} \right) \bO^\trans \by
        + \bO \underset{i}{\diag} \left( \frac{ \lambda_i^{(0)} }{ \gamma_t ( \lambda_i^{(0)} \gamma_t^- +1)} \right) \bO^\trans \by \\
        &= \bO \underset{i}{\diag} \left( \frac{ \gamma_{t-1}^- \lambda_i^{(0)}}{ \lambda_i^{(0)} \gamma_{t}^- + 1} +  \frac{ \lambda_i^{(0)} }{ \gamma_t ( \lambda_i^{(0)} \gamma_t^- +1)} \right) \bO^\trans \by \\
        &= \bO \underset{i}{\diag} \left(  \frac{ \gamma_t \gamma_{t-1}^- \lambda_i^{(0)} + \lambda_i^{(0)}}{ \gamma_t ( \lambda_i^{(0)} \gamma_t^- +1) }  \right) \bO^\trans \by \\
        &= \bO \underset{i}{\diag} \left(  \frac{ \lambda_i^{(0)} (\gamma_{t-1}^- +1/\gamma_t)}{  \lambda_i^{(0)} \gamma_t^- +1 }  \right) \bO^\trans \by \\
        &= \bO \underset{i}{\diag} \left(  \frac{ \lambda_i^{(0)} \gamma_{t}^- }{  \lambda_i^{(0)} \gamma_t^- +1 }  \right) \bO^\trans \by 
    \end{align*}
    as required.
\end{proof}

We may now prove the main result of this section.
\begin{theoremrep} \label{th:mainrec}
    The solution to the recursions \eqref{eq:recmean} and \eqref{eq:recvar} is given by
    \begin{align}
    m_t(x) &= k_{0}(x,\bx^\trans) ( \bK + \bI_N/\gamma_{t-1}^-)^{-1} \by \label{eq:recmeansol}\\
    k_t(x,y) &= k_0(x,y) - k_{0}(x,\bx^\trans) ( \bK + \bI_N/\gamma_{t-1}^-)^{-1} k_{0}(\bx,y) \nonumber 
    \end{align}
    for $t=1,2,\dots$.
\end{theoremrep}

\begin{proof}
    We first consider $k_t$. By combining \eqref{eq:recvar}, \eqref{eq:bKrekusion}, \eqref{eq:bKinvbK} and \eqref{eq:recfork}, we find
    \begin{align}
    k_{t+1}(x,y) =& k_t(x,y) - k_{0}(x,\bx^\trans) \bO \underset{i}{\diag} \left( \frac{1}{ \gamma_t ( \lambda_i^{(0)} \gamma_t^- +1)( \lambda_i^{(0)} \gamma_{t-1}^- +1)} \right) \bO^\trans k_{0}(\bx,y) \label{eq:tempexpkn} \\
    =& k_t(x,y) - k_{0}(x,\bx^\trans) \bO \underset{i}{\diag} \left( \frac{1}{ \lambda_i^{(0)}} \left( \frac{ 1 }{\lambda_i^{(0)} \gamma_{t-1}^- +1} - \frac{1}{\lambda_i^{(0)} \gamma_{t}^- +1} \right)\right) \bO^\trans k_{0}(\bx,y) \, . \nonumber
    \end{align}
    Iterating the equation above gives
    \begin{align*}
    k_{t}(x,y) &= k_{0}(x,y) - \sum_{j=0}^{n-1} k_{0}(x,\bx^\trans) \bO \underset{i}{\diag} \left( \frac{1}{ \lambda_i^{(0)}} \left( \frac{ 1 }{\lambda_i^{(0)} \gamma_{j-1}^- +1} - \frac{1}{\lambda_i^{(0)} \gamma_{j}^- +1} \right)\right) \bO^\trans k_{0}(\bx,y) \\
     &= k_{0}(x,y) - k_{0}(x,\bx^\trans) \bO   \underset{i}{\diag} \left( \frac{1}{ \lambda_i^{(0)}} \sum_{j=0}^{n-1}  \left( \frac{ 1 }{\lambda_i^{(0)} \gamma_{j-1}^- +1} - \frac{1}{\lambda_i^{(0)} \gamma_{j}^- +1} \right)\right) \bO^\trans k_{0}(\bx,y) \\
     &= k_{0}(x,y) - k_{0}(x,\bx^\trans) \bO   \underset{i}{\diag} \left( \frac{1}{ \lambda_i^{(0)}} \left( \frac{ 1 }{\lambda_i^{(0)} \gamma_{-1}^- +1} - \frac{1}{\lambda_i^{(0)} \gamma_{t-1}^- +1} \right)\right) \bO^\trans k_{0}(\bx,y) \\
    \end{align*}
    and since $\gamma_{-1}^- = 0$ we see that
    \begin{align*}
    k_{t}(x,y) &= k_{0}(x,y) - k_{0}(x,\bx^\trans) \bO   \underset{i}{\diag} \left( \frac{1}{ \lambda_i^{(0)}} \left( 1 - \frac{1}{\lambda_i^{(0)} \gamma_{t-1}^- +1} \right)\right) \bO^\trans k_{0}(\bx,y) \\
    &= k_{0}(x,y) - k_{0}(x,\bx^\trans) \bO   \underset{i}{\diag}  \left( \frac{1}{\lambda_i^{(0)} + 1/ \gamma_{t-1}^-} \right) \bO^\trans k_{0}(\bx,y) \\
    &= k_0(x,y) - k_{0}(x,\bx^\trans) ( \bK + \bI_N/\gamma_{t-1}^-)^{-1} k_{0}(\bx,y)
    \end{align*}
    as required.\\
    Next, we consider \eqref{eq:recmeansol}. From \eqref{eq:recmean} we have
    $$
    m_{t+1}(x) = m_t(x) + k_t(x,\bx^\trans) [\bK_t + \gamma_t \bI_N]^{-1} ( \by - m_t(\bx) )
    $$
    Using \eqref{eq:reckn}, \eqref{eq:recmn} and \eqref{eq:bKinvbK}, we may rewrite this as
    \begin{align}
    & m_{t+1}(x) = \nonumber\\
    & m_t(x) + k_0(x,\bx^\trans) \bO \underset{i}{\diag} \left( \frac{ 1 }{ \gamma_t ( \lambda_i^{(0)} \gamma_t^- +1)} \right) \bO^\trans (\by - \bO \underset{i}{\diag} \left( \frac{\lambda_i^{(0)}\gamma_{t-1}^- }{ \lambda_i^{(0)}\gamma_{t-1}^- + 1}  \right)\bO^\trans \by) \nonumber \\
    & m_t(x) + k_0(x,\bx^\trans) \bO \underset{i}{\diag} \left( \frac{ 1 }{ \gamma_t ( \lambda_i^{(0)} \gamma_t^- +1)} - \frac{ 1 }{ \gamma_t ( \lambda_i^{(0)} \gamma_t^- +1)}\frac{\lambda_i^{(0)}\gamma_{t-1}^- }{ (\lambda_i^{(0)}\gamma_{t-1}^- + 1)} \right) \bO^\trans \by \nonumber \\
    & m_t(x) + k_0(x,\bx^\trans) \bO \underset{i}{\diag} \left(  \frac{ 1 }{ \gamma_t ( \lambda_i^{(0)} \gamma_t^- +1)(\lambda_i^{(0)}\gamma_{t-1}^- + 1)} \right) \bO^\trans \by \label{eq:tempexpmn}
    \end{align}
    Notice, that the entries of the diagonal-matrix appearing in \eqref{eq:tempexpmn} are identical to those appearing in \eqref{eq:tempexpkn}. Using the same calculations as above, we find
    \begin{align*}
    m_{t+1}(x) &= m_0(x) + k_0(x,\bx^\trans)  \bO \underset{i}{\diag} \left( \frac{1}{\lambda_i^{(0)} +1/\gamma_{t-1}^-} \right) \bO^\trans \by \\
    &= k_0(x,\bx^\trans) \bO \underset{i}{\diag} \left( \frac{1}{\lambda_i^{(0)} +1/\gamma_{t-1}^-} \right) \bO^\trans \by \, ,
    \end{align*}
    from which \eqref{eq:recmeansol} follows.
\end{proof}

It follows from Theorem \ref{th:mainrec}, that performing multiple steps of distribution-centric self-distillation is equivalent to fitting an ordinary GP with a particular choice of noise parameter.  

Next, we note that if we assume that the noise parameter is identical in all distillation steps, i.e. $\gamma_t = \gamma$ for $t=0,1,\dots$, then $\gamma_{t-1}^- = t/\gamma$ which corresponds to fitting a GP to a dataset $\mathcal{D}_t$, which consists of $t$ replications of $\mathcal{D}$ with noise parameter $\gamma$, in a sense made precise by the following corollary.

\begin{corollaryrep}
\label{cor:datarep}
     If we let 
    $$
    \mathcal{D}_t = \{ (x_{ij}, y_{ij}) \mid i = 1,\dots,N \, , \quad j = 1,\dots, t \}
    $$
    where $x_{ij} = x_i$ for all $i$ and $j$, then the posterior distribution of $\bbf$ is $\mathcal{N}(\bmu_t,\bSigma_t)$ where
    \begin{align}
        \bmu_t &= \begin{pmatrix}
        & \bK(\bK + \gamma/t \bI_N)^{-1} \by \\
        & \vdots\\
        & \bK(\bK + \gamma/t \bI_N)^{-1} \by 
        \end{pmatrix} \\
        \bSigma_t &= \mathbf{1} \mathbf{1}^\trans \otimes (\bK - \bK(\bK + \gamma/t\bI_N )^{-1} \bK)
    \end{align}
    where $\otimes$ is the Kronecker product.
\end{corollaryrep}
\begin{proof}
    For simplicity, we only prove the statement for $t=2$, as the general proof is similar. The key observation is that the prior covariance matrix may be written as
    $$
    \begin{Bmatrix}
    \bK & \bK \\
    \bK & \bK 
    \end{Bmatrix} + \gamma\bI_{2 N}
    = 
    \begin{Bmatrix}
    \bK + \gamma\bI_N & \bK \\
    \bK & \bK + \gamma \bI_N
    \end{Bmatrix} 
    $$
    Using the formula for the inverse of a partioned matrix, we find
    $$
    \begin{Bmatrix}
    \bK + \gamma \bI_N & \bK \\
    \bK & \bK + \gamma \bI_N
    \end{Bmatrix}^{-1} =
    \begin{Bmatrix}
    \bA_{11} & \bA_{12} \\
    \bA_{21} & \bA_{22}
    \end{Bmatrix} 
    $$
    where
    \begin{align*}
        \bA_{11} &= \bA_{22} = (\bK + \gamma \bI_N - \bK(\bK + \gamma \bI_N)^{-1} \bK)^{-1} \\
        \bA_{12} &= \bA_{21} = - (\bK + \gamma \bI_N - \bK(\bK + \gamma \bI_N)^{-1} \bK)^{-1} \bK (\bK + \gamma \bI_N)^{-1}
    \end{align*}
    and these may be further simplified to obtain
    $$
    \left(
    \begin{Bmatrix}
    \bK & \bK \\
    \bK & \bK 
    \end{Bmatrix} + \bI_{2 N}
    \right)^{-1} =
    \begin{Bmatrix}
    - (2 \bK \gamma + \gamma^2 \bI_N)^{-1}(\bK + \gamma \bI_N) & (2 \bK \gamma + \gamma^2 \bI_N)^{-1} \bK \\
    - (2 \bK \gamma + \gamma^2 \bI_N)^{-1} \bK & (2 \bK \gamma + \gamma^2 \bI_N)^{-1}(\bK + \gamma \bI_N) 
    \end{Bmatrix} \, ,
    $$
    so we find
    \begin{align*}
    \bmu_2 &= 
    \begin{Bmatrix}
    \bK & \bK \\
    \bK & \bK 
    \end{Bmatrix}
    \begin{Bmatrix}
    (2 \bK \gamma + \gamma^2 \bI_N)^{-1}(\bK + \gamma \bI_N) & - (2 \bK \gamma + \gamma^2 \bI_N)^{-1} \bK \\
    - (2 \bK \gamma + \gamma^2 \bI_N)^{-1} \bK & (2 \bK \gamma + \gamma^2 \bI_N)^{-1}(\bK + \gamma \bI_N)
    \end{Bmatrix}
    \begin{pmatrix}
    \by \\
    \by
    \end{pmatrix} \\
    &= 
    \begin{Bmatrix}
    \bK & \bK \\
    \bK & \bK 
    \end{Bmatrix}
    \begin{pmatrix}
    (2 \bK \gamma + \gamma^2 \bI_N)^{-1} \gamma \by \\
    (2 \bK \gamma + \gamma^2 \bI_N)^{-1} \gamma \by 
    \end{pmatrix}
    \\
    &= 
    \begin{pmatrix}
    2 \bK (2 \bK \gamma + \gamma^2 \bI_N)^{-1} \gamma \by \\
    2 \bK (2 \bK \gamma + \gamma^2 \bI_N)^{-1} \gamma \by 
    \end{pmatrix}
    = 
    \begin{pmatrix}
    \bK ( \bK  + (\gamma/2) \bI_N)^{-1}  \by \\
    \bK ( \bK  + (\gamma/2) \bI_N)^{-1}  \by 
    \end{pmatrix} \, .
    \end{align*}
    Similarly, the covariance matrix is found as
    \begin{align*}
    \bSigma_2 =
    \begin{Bmatrix}
    \bK & \bK \\
    \bK & \bK 
    \end{Bmatrix}
    -
    \begin{Bmatrix}
    \bK & \bK \\
    \bK & \bK 
    \end{Bmatrix}
    \begin{Bmatrix}
    (2 \bK \gamma + \gamma^2 \bI_N)^{-1}(\bK + \gamma \bI_N) & - (2 \bK \gamma + \gamma^2 \bI_N)^{-1} \bK \\
    - (2 \bK \gamma + \gamma^2 \bI_N)^{-1} \bK & (2 \bK \gamma + \gamma^2 \bI_N)^{-1}(\bK + \gamma \bI_N)
    \end{Bmatrix}
    \begin{Bmatrix}
    \bK & \bK \\
    \bK & \bK 
    \end{Bmatrix}
    \end{align*}
    which may be simplified to give
    $$
    \bSigma_2 = 
    \begin{Bmatrix}
    \bK - \bK ( \bK  + (\gamma/2) \bI_N)^{-1} \bK & \bK - \bK ( \bK  + (\gamma/2) \bI_N)^{-1} \bK \\
    \bK - \bK ( \bK  + (\gamma/2) \bI_N)^{-1} \bK & \bK - \bK ( \bK  + (\gamma/2) \bI_N)^{-1} \bK 
    \end{Bmatrix} \, ,
    $$
    as desired.
\end{proof}

\subsection{Illustratory Example}
We now consider an illustratory example, where we assume a true function $g(z) = z\sin{(z)}$, 10 training samples $\bx$ equidistantly distributed between 0 and 10, and observed values $y_i = g(x_i) + \varepsilon_i$, where $\varepsilon_i$ is standard normally distributed. We use a scaled radial basis kernel with two hyperparameters $\sigma_f$ and $l$ as well as noise parameters $\gamma_t$. See Section \ref{sec:example_details} for further details on the setup.

We now consider $10$ steps of both data-centric and distribution-centric self-distillation with fixed hyperparameters for all steps. In Figure \ref{fig:gpr_data_example} and Figure \ref{fig:gpr_distribution_example} we plot the results for data-centric distillation with $(\gamma_1, \dots, \gamma_{10}) = (0.1, \dots, 1)$. Here, we observe the expected increase in regularization of the mean, and that the uncertainty increases with each step as expected from the choice of $(\gamma_1, \dots, \gamma_{10})$. This aligns well with our theory. However, for the distribution-centric case, the mean function is nearly unaffected by the distillation procedure, and the uncertainty estimate does not change much either. This is due to the relatively small change in $1/\gamma_{t}^{-}$ when $t$ goes from $1$ to $10$ as  $1/\gamma_{1}^{-} = 0.1$ and $1/\gamma_{10}^{-} = 0.034$, which yield a small change for each distillation step. See Appendix \ref{sec:example_details} for additional examples with different choices of parameters.

\section{Self-Distillation for GP Classification}\label{sec:gpc_sd}
In the following we separately investigate data-centric and distribution-centric self-distillation for GP classification.

\subsection{Data-centric Self-Distillation}\label{sec:data_gpc}

\begin{figure*}[htbp]
    \centering
    \begin{minipage}{.49\linewidth}
        \centering
        \includegraphics[width=\linewidth]{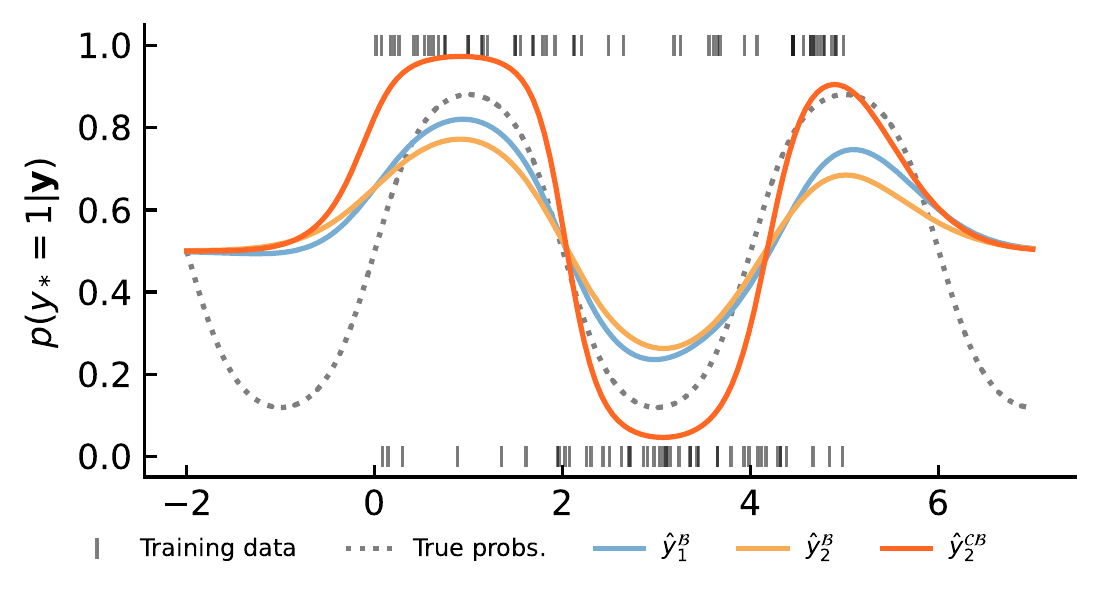}
        \caption{Ordinary GPC model and one step of data-centric distillation, both with continuous Bernoulli and ordinary Bernoulli likelihood. We note that the Bernoulli likelihood, although, well-performing does not constitute a well-specified model.}
        \label{fig:GPC_data_example_cb_vs_b}
    \end{minipage}
    \hfill
    \begin{minipage}{.49\linewidth}
        \centering
        \includegraphics[width=\linewidth]{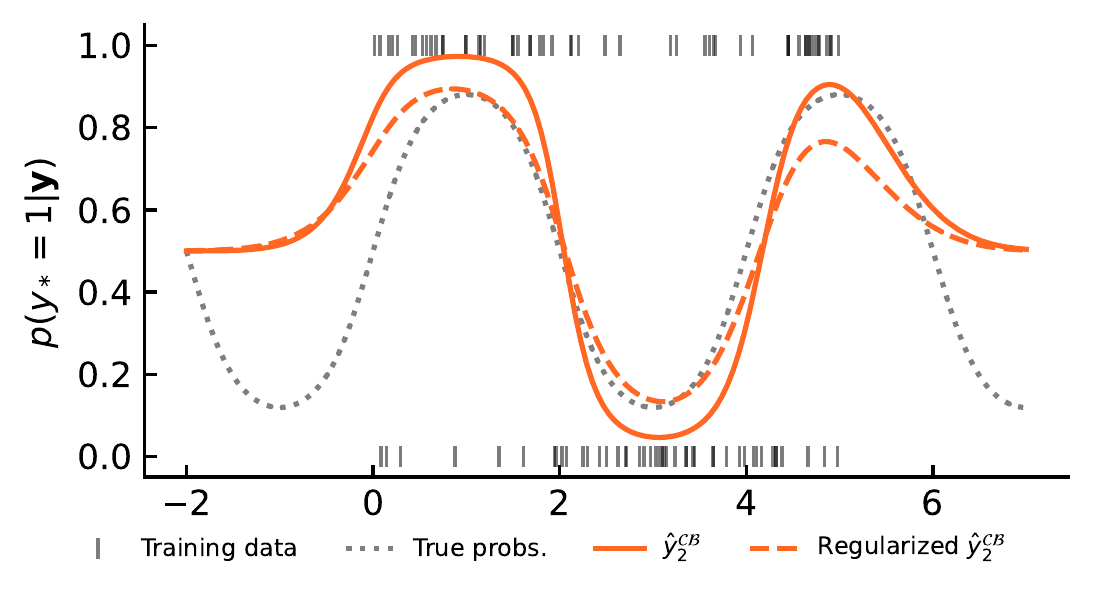}
        \caption{One step of data-centric distillation with continuous Bernoulli likelihood as in Figure \ref{fig:GPC_data_example_cb_vs_b}, both with and without regularization imposed by a noise assumption on the training observations. Adjusting the noise parameter $\gamma_1$ adjust the regularization.}
        \label{fig:GPC_data_example_regularized_cb}
    \end{minipage}
\end{figure*}
Similarly to the data-centric approach for GPR, we will in this section consider the posterior mean predictions as deterministic targets for the next iteration of the model. In particular, at each step, $t$, we assume the prior of $f_{t}$ to be $\mathcal{GP}(0,k)$, and for $t=1$ we will obtain predictions, $\hat{\by}_{0}$ as the average posterior prediction based on the Laplace approximation e.g. $\hat{\by}_{0} = \E_q[\sigma(\bbf_{1}) \mid \by_{0}]$. These predictions will then be used in place of the original observations for step $t=2$, which will lead to new predictions $\hat{\by}_{2}$, that we can use in the succeeding step, and so on.

Since the original observations, $\by_{0}$, are in $\{0,1\}^N$ and the predictions, $\hat{\by}_{1}$, are in $[0,1]^N$, directly reusing the predictions, $\hat{\by}_{1}$, in place of the observations, $\by_{0}$, in the subsequent step yields a misspecified model as the Bernoulli distribution only has support on $\{0,1\}$. To alleviate this misspecification, we redefine our model for $t \geq 2$ to assume a conditional \textit{continuous} Bernoulli distribution rather than the usual conditional Bernoulli \citep{Loaiza-Ganem2019TheAutoencoders}. We will occasionally refer to this method as C-GPC, and in practice, this requires that we multiply each element of \eqref{eq:GPdef} by the appropriate normalizing constant, $C$ (see \eqref{eq:c_constant} and Appendix \ref{sec:cont_bern_details}). That is, for $t \geq 2$ we assume
\begin{align*}
    &\by_{t-1} \mid \bbf_{t} \sim \prod_{n=1}^N C(\sigma(f_{t,n}))\sigma(f_{t,n})^{y_{t-1,n}}(1-\sigma(f_{t,n}))^{1-y_{t-1,n}},
\end{align*}
which in turn affects the log posterior $\log \tilde{p}(\bbf_{t} \mid \by_{t-1})$ through $\log \tilde{p}(\by_{t-1} \mid \bbf_{t})$ which now becomes
\begin{align*}
    \log \tilde{p}(\by_{t-1} \mid \bbf_{t})
    &= \log p(\by_{t-1} \mid \bbf_{t}) + \sum_{n=1}^N \log C(\sigma(f_{t,n})).
\end{align*}
Furthermore, this also affects the gradient and hessian of $\log \tilde{p}(\bbf_{t} \mid \by_{t-1})$ used to obtain the Laplace approximation, and although $\log C(\lambda)$ nor $\frac{d}{d\lambda}\log C(\lambda)$ attains no simple expressions, we show in Proposition \ref{prop:C_derivatives} that both $C(\sigma(a))$ and the derivatives $\frac{d}{da} \log C(\sigma(a))$ and $\frac{d^2}{d^2a} \log C(\sigma(a))$ yield surprisingly simple expressions of standard hyperbolic functions (see also Figure \ref{fig:cont_bernoulli_constant} for plots of the functions).
\begin{proprep}\label{prop:C_derivatives}
Let $C(\sigma(a))$ be defined as in \eqref{eq:c_constant} with $\lambda = \sigma(a)$, then we have that
    \begin{align*}
        C(\sigma(a)) &= \begin{cases}
            2 & \text{if } a = 0 \\
            a\mathrm{coth}\left(\frac{a}{2}\right) & \text{otherwise},
        \end{cases} \\
        \frac{d}{da} \log(C(\sigma(a))) &= \begin{cases}
            0 & \text{if } a = 0 \\
            \frac{1}{a} - \frac{1}{\mathrm{sinh}(a)} & \text{otherwise},
        \end{cases}  \\
        \frac{d^2}{d^2a} \log(C(\sigma(a))) &= \begin{cases}
            \frac{1}{6} & \text{if } a = 0 \\
            -\frac{1}{a^2} + \frac{\mathrm{coth}(a)}{\mathrm{sinh}(a)} & \text{otherwise},
        \end{cases}
    \end{align*}
    where $\mathrm{coth}$, and $\mathrm{sinh}$ are the hyperbolic cotangent, and sine functions, respectively.
\end{proprep}
\begin{proof}
  First we note that for $a \neq 0$ it can be shown that
  \begin{align}\label{eq:c_sigma_partial}
      C(\sigma(a)) &= \frac{2\mathrm{tanh}^{-1}(1-2\sigma(a))}{1-2\sigma(a)} = a\mathrm{coth}\left(\frac{a}{2}\right),
  \end{align}
  which follows from the fact that $\mathrm{tanh}(x) = \frac{e^{2x}-1}{e^{2x}+1}$
  \begin{align*}
      1-2\sigma(a) = 1 - \frac{2e^{a}}{1+e^{a}} = \frac{1 + e^{a}}{1 + e^{a}} - \frac{2e^{a}}{1+e^{a}} = \frac{1 - e^{a}}{1 + e^{a}} = \mathrm{tanh}\left(-\frac{a}{2}\right)
  \end{align*}
  and thus it follows directly that
  \begin{align*}
      2\mathrm{tanh}^{-1}(1-2\sigma(a)) = 2 \mathrm{tanh}^{-1}\left( \mathrm{tanh}\left(-\frac{a}{2}\right)\right) = \frac{-2a}{2} = -a
  \end{align*}
  and finally by $\mathrm{coth}(x) = \frac{1}{\mathrm{tanh}(x)}$ and $\mathrm{tanh}(-x) = -\mathrm{tanh}(x)$ that
  \begin{align*}
      \frac{1}{1 - 2\sigma(a)} = \frac{1}{-\mathrm{tanh}\left(\frac{a}{2}\right)} = - \mathrm{coth}\left(\frac{a}{2}\right),
  \end{align*}
  which combines to the claim in \eqref{eq:c_sigma_partial}.
  Now, By properties of the hyperbolic functions\footnote{In particular that $\mathrm{sinh}(2x) = 2\mathrm{sinh}(x)\mathrm{cosh}(x)$ and that $\frac{d}{da}\mathrm{coth}(x) = -\frac{1}{\mathrm{sinh}^2(x)}$ for $x \neq 0$.} we get by direct computation that
  \begin{align*}
      \frac{d}{da} \log(C(\sigma(a))) &= \frac{d}{da} \log(a) + \frac{d}{da} \log\mathrm{coth}\left(\frac{a}{2}\right)  & \text{for } a \neq 0\\
      &= \frac{1}{a} + \frac{1}{\mathrm{coth}\left(\frac{a}{2}\right)}\frac{d}{da}\mathrm{coth}\left(\frac{a}{2}\right) & \text{for } a \neq 0\\
      &= \frac{1}{a} + \frac{\mathrm{sinh}\left(\frac{a}{2}\right) }{\mathrm{cosh}\left(\frac{a}{2}\right)}\left(- \frac{1}{2\mathrm{sinh}^2\left(\frac{a}{2}\right)}\right) & \text{for } a \neq 0\\
      &= \frac{1}{a} - \frac{1}{2\mathrm{cosh}\left(\frac{a}{2}\right)\mathrm{sinh}\left(\frac{a}{2}\right)} & \text{for } a \neq 0\\
      &= \frac{1}{a} - \frac{1}{\mathrm{sinh}(a)} \qquad {\color{gray}\left(= \frac{1}{a} - \mathrm{csch}(a)\right)} & \text{for } a \neq 0.
  \end{align*}
  Additionally, by further properties of the hyperbolic functions\footnote{Here, in particular $\frac{d}{da}\mathrm{csch}(x) = - \mathrm{csch}(x)\mathrm{coth}(x)$, $\mathrm{coth}(x) = \frac{1}{\mathrm{tanh}(x)}$ and $\mathrm{csch}(x) = \frac{1}{\mathrm{sinh}(x)}$.} it follows directly that
  \begin{align*}
      \frac{d^2}{d^2a} \log(C(\sigma(a))) &= -\frac{1}{a^2} - \frac{d}{da}\mathrm{csch}(x) & \text{for } a \neq 0 \\
      &= -\frac{1}{a^2} + \mathrm{csch}(a)\mathrm{coth}(a) &  \text{for } a \neq 0. \\
      &= -\frac{1}{a^2} + \frac{1}{\mathrm{sinh}(a)}\frac{1}{\mathrm{tanh}(a)} &  \text{for } a \neq 0.
  \end{align*}
  It remains to show both derivatives for $a = 0$.
  First let $f(a) = \log(C(\sigma(a))) = \log a\mathrm{coth}\left(\frac{a}{2}\right)$ for $a \neq 0$ and $\log(2)$ elsewhere, then we have that $f(-x) = \log \left(-x\mathrm{coth}\left(\frac{-x}{2}\right)\right) = \log\left(x\mathrm{coth}\left(\frac{x}{2}\right)\right) = f(x)$, and we can without loss of generality merely consider limits from the right. See that
  \begin{align*}
      \lim_{h \downarrow 0} \frac{f(h) - f(0)}{h} &= \lim_{h \downarrow 0} \frac{d}{dx}f(h) \\
      &= \lim_{h \downarrow 0} \left(\frac{1}{h} - \frac{1}{\mathrm{sinh}(h)}\right) \\
      &= \lim_{h \downarrow 0} \left(\frac{\mathrm{sinh}(h)-h}{h\mathrm{sinh}(h)}\right),
  \end{align*}
  and by applying L'Hôpital's rule for $0/0$ limits we have that
    \begin{align*}
      \lim_{h \downarrow 0} \frac{f(h) - f(0)}{h} &= \lim_{h \downarrow 0} \left(\frac{\mathrm{cosh}(h)-1}{\mathrm{sinh}(h) + x\mathrm{cosh}(h)}\right) = 0.
  \end{align*}
  By a similar argument we get that since $\mathrm{sinh}(-x) = -\mathrm{sinh}(x)$ then $\frac{d}{dx} f(-x) = -\frac{d}{dx}f(x)$, and we can w.l.o.g. look only at right limits and apply L'Hôpital's rule, which yields that
  \begin{align*}
      \lim_{h \downarrow 0} \frac{\frac{d}{dx}f(h) - \frac{d}{dx}f(0)}{h} &= \lim_{h \downarrow 0} \frac{d^2}{d^2x}f(h) = \lim_{h \downarrow 0} \left(-\frac{1}{a^2} + \frac{1}{\mathrm{sinh}(a)}\frac{1}{\mathrm{tanh}(a)} \right) = \frac{1}{6}
  \end{align*}
  and we are done.
\end{proof}
Thus, if we denote $\bC_{t} \eqdef \sum_{n=1}^N \log(C(\sigma(f_{t,n})))$, we get that the gradient and hessian of the log posterior becomes
\begin{align*}
    \nabla \tilde{\psi}( \bbf_{t}) &= \by_{t-1} - \bsigma_{t-1} - \bK^{-1} \bbf_{t}  + \nabla\bC_{t}, \\
    -\nabla \nabla \tilde{\psi}( \bbf_{t}) &= \bW_{t} + \bK^{-1} - \nabla\nabla\bC_{t},
\end{align*}
where $\bW_{t} = \diag_n \{\sigma(f_{t,n})(1-\sigma(f_{t,n}))\}$ and $\nabla\nabla\bC_{t} = \diag_n\{\frac{d^2}{d^2f_{t,n}}\log C(\sigma(f_{t,n}))\}$, respectively. These directly affect the Laplace approximation of the posterior as is also evident from e.g. Figure \ref{fig:GPC_data_example_cb_vs_b}.

Unfortunately, unlike data-centric self-distillation for GPR, we are unable to obtain a closed-form solution for the distilled models due to the numerical approximations necessary to obtain the posterior (predictions). However, we do empirically investigate the effect of using the continuous Bernoulli rather than the discrete Bernoulli distribution in Figure \ref{fig:GPC_data_example_cb_vs_b}. In particular, we plot the solutions for a single step of distillation with C-GPC and with the discrete Bernoulli where we use continuous observations despite the support of the Bernoulli distribution being discrete. We expand on this example in Section \ref{sec:gpc_example} below.

\subsection{Distribution-centric Self-Distillation}\label{sec:distribution_gpc}
\begin{figure*}[htbp]
    \centering
    \begin{subfigure}[b]{0.48\linewidth}
        \centering
        \includegraphics[width=\linewidth]{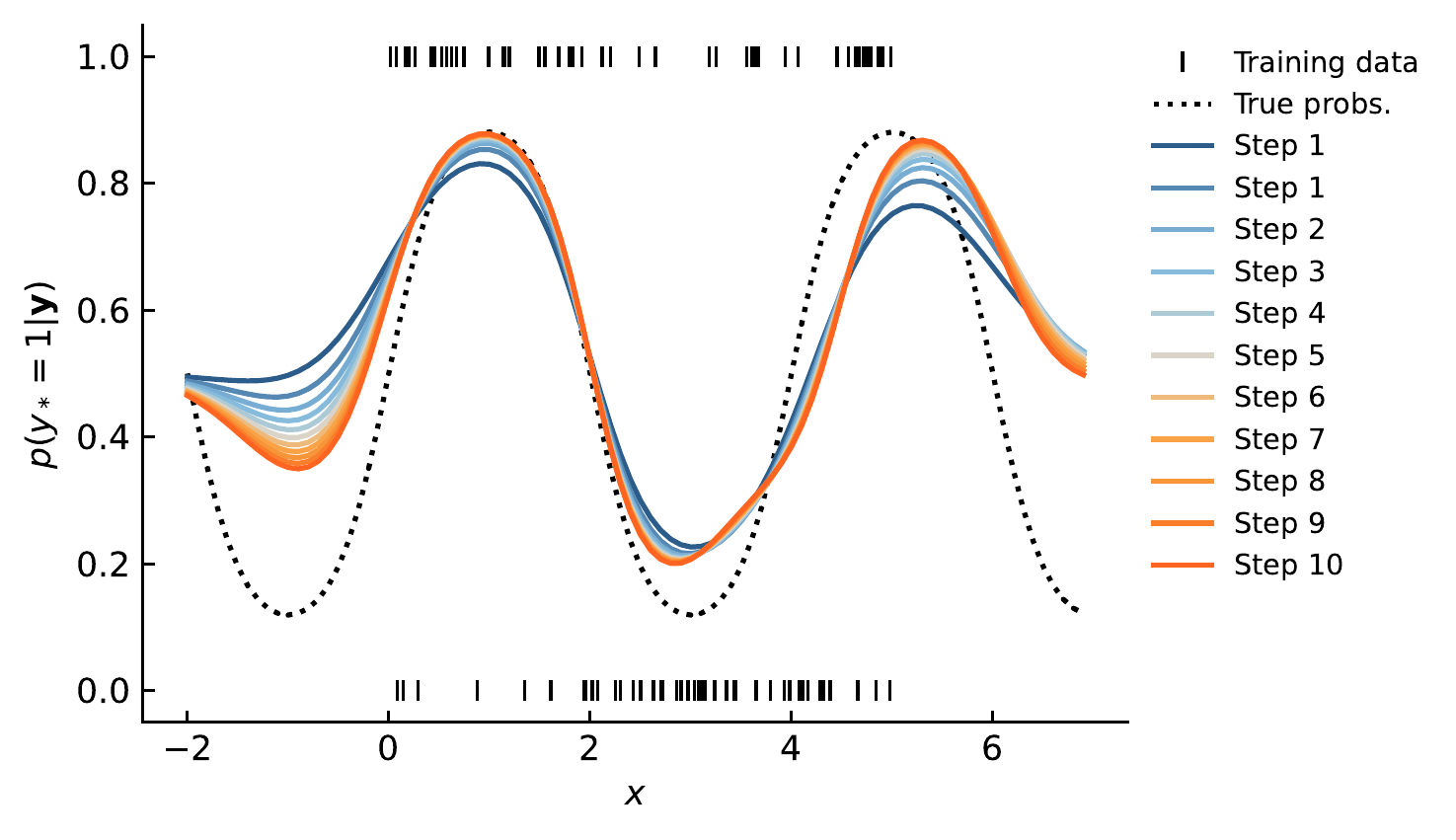}
        \caption{Solutions for 10 steps of distillation.}
        \label{fig:gpc_distribution_example_fit}
    \end{subfigure}
    \hfill
    \begin{subfigure}[b]{0.48\linewidth}
        \centering
        \includegraphics[width=\linewidth]{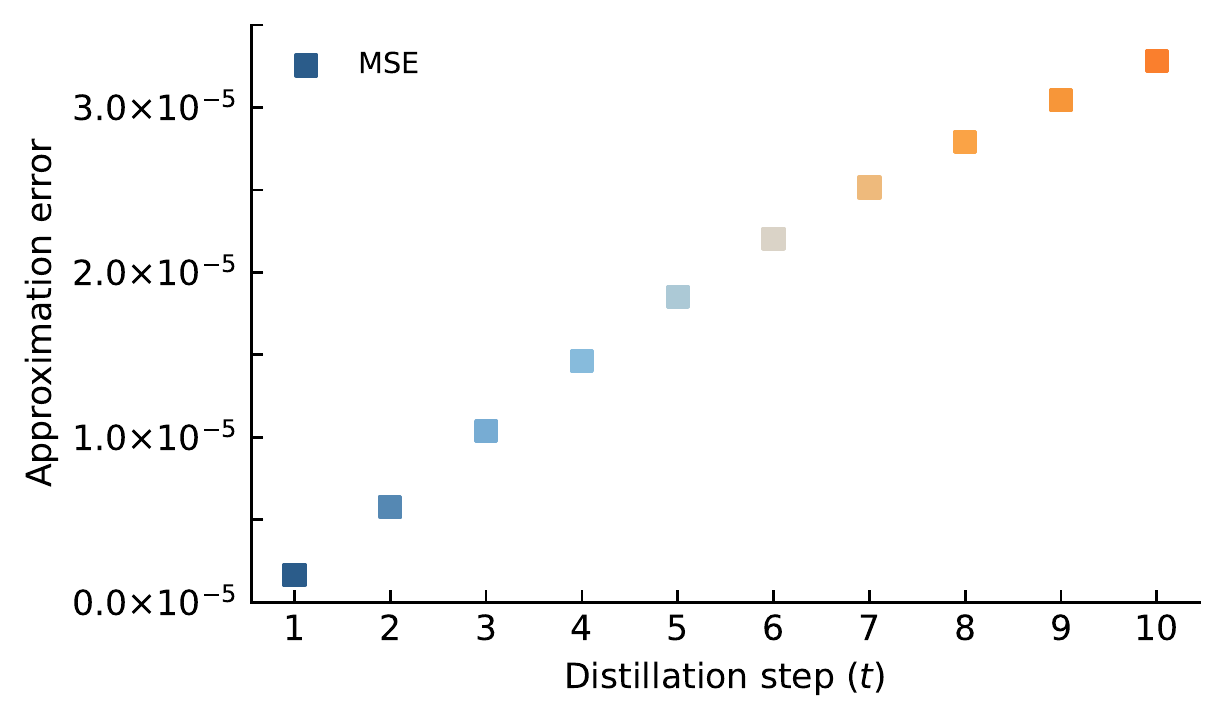}
        \caption{Approximation error between iterated and scaled solutions.}
        \label{fig:gpc_distribution_example_error}
    \end{subfigure}
    \caption{Ten steps of distribution-centric self-distillation for GP classification. In (\protect\subref{fig:gpc_distribution_example_fit}) we plot the solutions using the scaled covariance approximation, and in (\protect\subref{fig:gpc_distribution_example_error}) we plot the approximation error between the iterated distillation procedure and scaled approximation over the test set of 90 equidistant point on the interval $[-2,7]$.}
    \label{fig:XXX}
\end{figure*}
In the distribution-centric approach, we proceed in a similar manner as in Section \ref{sec:distribution_sd}. Firstly, we observe from 
\eqref{eq:gppredmean} and \eqref{eq:gppredbSigma} that our approximate posterior distribution is a Gaussian process 
$\mathcal{GP}(m_\ast,k_\ast)$ with
\begin{align*}
    m_\ast(x) &= m(\bx) + k(\bx, \bx^\trans) \bK^{-1} (\hat{\bbf} - m(\bx)) \\
    k_\ast(x,y ) &= k(x, y) - k(x, \bx^\trans) (\bK + \bW^{-1})^{-1}k(\bx, y) \, .
\end{align*}
This may then be used as a prior distribution, 
which again implies a series of Gaussian processes $\mathcal{GP}(m_t,k_t)$, $t = 0,1,2,\dots$ where analogue to equations 
\eqref{eq:recmean} and \eqref{eq:recvar} in this case becomes
\begin{align}
    m_{t+1}(x)  &= m_t(x) +  k_t(x,\bx^\trans) \bK_t^{-1}( \hat{\bbf}_t - m_t(\bx)) \\
    k_{t+1}(x,y) &= k_t(x,y) -  k_t(x,\bx^\trans) (\bK_t + \bW^{-1}_t)^{-1} k_t(\bx,y). 
\end{align}
Here, $\hat{\bbf}_t$ is the mode of $\psi_t(\bbf)$, the log posterior for the $t$'th iteration, and similarly $\bW_t = \diag_n \{ \sigma(\hat{f}_{t,n}) (1 - \sigma(\hat{f}_{t,n}))\}$.

Since $\hat{\bbf}_t$ is not given analytically, we cannot solve the above recursions in the classification case, as we could in the regression setting and must resort to numerical results. However, we show an analogous result to the data duplication result from Corollary \ref{cor:datarep} and it follows that iterated distillation in the distribution-centric setup may be approximated by scaling the covariance in the first distillation step. These observations are summarized in the following proposition, where $\mathcal{D}_t$ as
 in Corollary \ref{cor:datarep} consists of the dataset $\mathcal{D}$ replicated $t$ times, and an example is illustrated in Figure \ref{fig:gpc_distribution_example_fit} and \ref{fig:gpc_distribution_example_error}.
\begin{proprep} \label{prop:gpc_scaling}
$(\bA)$ A single step of distribution-centric distillation using $\mathcal{D}_t$ and a Gaussian prior $\mathcal{GP}(0,k)$ yields the same posterior distribution as a single step of distribution-centric distillation using a Gaussian prior $\mathcal{GP}(0,t k)$. $(\bB)$ Performing $t$ iterations of the distribution-centric distillation using $\mathcal{D}$ and starting with an inital Gaussian prior $\mathcal{GP}(0,k)$ yields \textit{approximately} the same posterior as a single step of distribution-centric distillation using a Gaussian prior $\mathcal{GP}(0,t k)$.
\end{proprep}

\begin{proof}
$(\bA)$ \\
First, we notice that duplicating the data simply implies that the term corresponding to the conditional distribution of the data given the latent variables is repeated $t$ times. Using the notation from section \ref{sec:GPC}, this implies that the log posterior will satisfy 
$$
\psi( \bbf) \eqdef \log p(\bbf \mid \by) \underset{\bbf}{\propto} \log p(\bbf) + t \log p(\by \mid \bbf)
$$
with gradient
$$
\nabla \psi( \bbf) = - \bK^{-1} \bbf   + t (\by - \bsigma ) 
$$
and by setting this equal to zero, we see that the maximizer $\hat{\bbf}$ will satisfy
$$
\hat{\bbf} = (t \bK) (\by - \bsigma ) \, .
$$
Comparing to the same equation for single-step distillation implied by \eqref{eq:optimlogit}, we see that the two methods give the same result. \\
Next, we wish to show $(\bB)$. For simplicity, we consider $t = 2$ as the argument for general $t$ is similar. From $(\bA)$ we see that we can obtain the desired result by showing that $2$ steps of distillation will give approximately the same result as a single step of distillation based on $\mathcal{D}_2$. For the latter, we see from the proof of $(\bA)$ that the posterior distribution is proportional to
\begin{align}
    \left( \prod_{n=1}^N \sigma(f_n)^{t_n}(1-\sigma(f_n))^{1-t_n} \right)^2 \exp\left(-\frac{1}{2} {\bbf^\trans \bK^{-1} \bbf} \right) \, . \label{eq:twostepdistill}
\end{align}
If we define
$$
\varphi(\bbf) \eqdef \left( \prod_{n=1}^N \sigma(f_n)^{t_n}(1-\sigma(f_n))^{1-t_n} \right) \exp\left(-\frac{1}{2} {\bbf^\trans \bK^{-1} \bbf} \right)
$$
we may rewrite \eqref{eq:twostepdistill} as
\begin{align}
\left( \prod_{n=1}^N \sigma(f_n)^{t_n}(1-\sigma(f_n))^{1-t_n} \right)
\varphi(\bbf) \, . \label{eq:twostepdistill2}
\end{align}
Now we simply observe that $2$-step distillation will correspond to replacing $\varphi(\bbf)$ in \eqref{eq:twostepdistill2} with the 
appropriate Gaussian distribution coming from the Laplace distribution, and hence the approximation comes from the fact that $\varphi(\bbf)$ is not itself exactly proportional to a Gaussian distribution.
\end{proof}

\subsection{Illustratory Example}\label{sec:gpc_example}
We now consider an illustratory example, where we consider the true underlying function $g(x) = 2\mathrm{sin}\left(\frac{x\pi}{2}\right)$. We sample training observations, $\bx$, from a $\mathcal{U}(0,5)$ distribution and binary targets from a Bernoulli distribution with probability parameterized by $g(x)$.

\paragraph{Data-centric GPC distillation}
For data-centric distillation, we initially fit an ordinary GP classification model with Bernoulli distribution to the binary observations and fit a secondary model to the continuous predictions of this first model. In Figure \ref{fig:GPC_data_example_cb_vs_b} we plot the predictions of the initial ordinary GPC, along with the predictions of the secondary model, when we use either a continuous Bernoulli distribution (as argued in Section \ref{sec:gpc_sd}) or a discrete Bernoulli distribution where we intentionally violate the discrete support of the distribution. The hyperparameters of the models are chosen to minimize the negative log-likelihood (NLL) over the input data, and while the distilled Bernoulli model is slightly flattened, yet well-performing, compared to the initial fit, the C-GPC model yields more ''extreme'' predictions. We conjecture that the implicit regularization imposed by the misspecification of the distilled Bernoulli model is absent in the C-GPC model. Furthermore, by Section \ref{sec:cont_bern_details} we expect the continuous Bernoulli to favor more extreme values compared to the ordinary Bernoulli, as is also evident in Figure \ref{fig:GPC_data_example_cb_vs_b}. To dampen the degree of overfitting in the C-GPC model we reintroduce the noise parameter $\gamma_t > 0$ along the diagonal of the kernel matrix of the training data. We plot both the regularized and non-regularized solutions in Figure \ref{fig:GPC_data_example_regularized_cb}, and observe that with the noise parameter, we get not only a well-specified model, but also get more control of the degree of regularization. See Appendix \ref{sec:example_details} for more experiments.

\paragraph{Distribution-centric GPC distillation}
Similarly to the data-centric case, we initially fit an ordinary GPC, but unlike the data-centric case, for distribution-centric distillation, we do not change the model assumptions for the distillation step. In Figure \ref{fig:gpc_distribution_example_fit} we plot the solutions for 10 steps of distillation and observe a significantly different behavior compared to data-centric distillation; the effect of distillation is much more subtle and more closely resembles that of distillation in the regression case and the solutions progressively fit closer and closer to the original data. Furthermore, in Figure \ref{fig:gpc_distribution_example_error} we plot the mean squared approximation error between the solutions obtained from \eqref{eq:recmean}-\eqref{eq:recvar} and those obtained by scaling the kernel matrix as argued in Proposition \ref{prop:gpc_scaling}. Although the error is almost linearly increasing, it does so by less than $0.5 \times 10^{-5}$ per step, and the approximation error is on the order of $10^{-5}$ for 10 steps. Thus, since the scaling approximation is much faster (especially for many distillation steps) it is a great alternative for naïve computations.

\section{Related Works}\label{sec:related_works}
To the best of our knowledge, no literature on knowledge distillation for Gaussian process models exists. However, for good measure in the following we relate our proposed methods to the existing related literature on knowledge distillation and Gaussian process regression/classification.

\paragraph{Knowledge Distillation}
The idea of knowledge distillation originates back to \citet{ba2013deep, Bucil2015}, but is typically most known from \citet{hinton2015distilling}. It was originally aimed at model compression, by training a small (in an appropriate measure of model size) \textit{student} model on the outputs of a larger \textit{teacher} model.

In recent years, the idea of distillation has expanded in numerous directions, and for various types of applications. Of particular relevance is the direction of self-distillation, where the student model is considered to be of the same model class/architecture as the teacher model \citep{furlanello2018born, Zhang2019BeDistillation, Allen-Zhu2020}. Other interesting directions of research on distillation are on the impact of the size-gap between teacher and student \citep{mirzadeh2019improved}, matching on other information than the output \citep{park2019relational, romero2014fitnets, srinivas2018knowledge}, as well as multi-teacher and multi-source distillation \citep{Li2021, Liu2020, 10.1145/3097983.3098135, AnonymousAuthors2023LearningTasks}. Finally, due to the domain-agnostic nature of most distillation methods, they are applied in a wide range of domains \citep{KrishnanParthasarathi2019LessonsSpeech, Borup2023AutomaticData}.

Despite much empirical success of a wide array of distillation techniques, a rigorous understanding of distillation is still lacking behind. However, recent work has provided insights into simplified settings such as linear models, kernel ridge regression, and decision trees \citep{phuong2019towards, Borup2021, Frosst2018DistillingTree}.

\paragraph{Gaussian Processes}
Many results on Gaussian Process models have been well-known for years and are collected in foundational books such as \citet{Rasmussen2006, bishop2007pattern, murphy2013machine}. However, in recent years, an increasing amount of research on extending the capabilities of GP models and adapting them to modern compute availability and dataset sizes has been made. This includes theoretical connections between neural networks and Gaussian processes \citep{Lee2018, Yang2019, Garriga-Alonso2019DeepProcesses}, stochastic and sparse GPs \citep{Titsias, Hensman2013GaussianData, Hensman2015}, and the possibility to exploit the successes of neural networks in combination with Gaussian processes e.g. deep kernel learning \citep{Wilson2016, Wilson2016StochasticLearning}.

\section{Conclusion}
In conclusion, we propose two approaches to extend knowledge distillation to Gaussian process regression and classification; data-centric and distribution-centric. We show that some of these approaches are closely related to known results or very particular choices of hyperparameters, but also that some methods require careful redefinition of our model assumptions to be appropriate. To the best of our knowledge, this paper is the first attempt to propose knowledge distillation specifically for Gaussian Process models, opening new avenues for further research with potential implications in machine learning. Interesting directions of future research include, amongst others, utilization of a convex combination of both the original and predicted targets in the distillation steps, as well as combinations of data-centric and distribution-centric distillation approaches.

\bibliography{arxiv}
\bibliographystyle{icml2023}

\newpage
\appendix
\section{Additional details on Self-distillation for GP classification}\label{sec:data_gpc_details}
In the following, we restate some known results for GP classification with more details than in the main paper. We also provide some additional details on our self-distillation results for GP classification.

\subsection{Gaussian Process Classification}
\label{sec:GPC}
Following the usual setup for classification with Gaussian Processes, we assume $f \sim \mathcal{GP}(m,k)$, and 
\begin{align*}
    \by \mid \bbf \sim \prod_{n=1}^N \sigma(f_n)^{t_n}(1-\sigma(f_n))^{1-t_n},
\end{align*}
and denote $\by = (y_1, \dots, y_n)^\trans$ with $y_i \in \{0,1\}$ and similarly $\bbf$ the vector of realizations of the GP-prior at the respective $x_i$ values. Then we have the log density
\begin{align*}
    \psi(\bbf) \eqdef \log p(\bbf \mid \by) \propto \log p(\bbf) + \log p(\by \mid \bbf),
\end{align*}
where $\log p(\bbf)$ is a multidimensional Gaussian, and one can show that
\begin{align*}
    \log p(\by \mid \bbf) = \by^\trans \bbf - \sum_{n=1}^N \log (1+\exp(f_n)) \, ,
\end{align*}
so that (up to the proportionality in $\bbf$):
\begin{align}
\psi( \bbf) = - \frac{1}{2} (\bbf_N-\bbm)^\trans \bK^{-1} (\bbf-\bbm) + \by^\trans \bbf - \sum_{n=1}^N \log (1+\exp(f_n)) \, .
\label{eq:fulllogpost}
\end{align}
Since the posterior is non-Gaussian, we apply Laplace approximation to get a Gaussian approximation.

Thus, we need the gradient and Hessian to obtain $\hat{\bbf}$ (the MAP estimate) and the covariance matrix. It follows from \eqref{eq:fulllogpost} that
\begin{align*}
    \nabla \psi( \bbf) &= \by - \bsigma - \bK^{-1} \bbf  + \bK^{-1}\bbm \\
    -\nabla \nabla \psi( \bbf) &= \bW + \bK^{-1}
\end{align*}
where $(\bsigma)_n = \sigma(f_n)$ (this term comes from $\sum_{n=1}^N \log( 1 + \e^{f_n} )$) and $\bW = \diag_n \{ \sigma(f_n) (1 - \sigma(f_n)) \}$. 

Using these can use the Newton-Raphson procedure to find $\hat{\bbf}$, and it follows from the expression for the gradient, that $\hat{\bbf}$ will be the solution to
\begin{align}
\hat{\bbf} = \bbm + \bK (\by - \bsigma) \, .
\label{eq:optimlogit}
\end{align}
The Gaussian approximation of $p(\bbf \mid \by)$ comes out to
$$
q(\bbf) = \mathcal{N}( \bbf \mid \hat{\bbf},   (\bW + \bK^{-1})^{-1} ) 
$$
The posterior predictive distribution is computed as
\begin{align}
p(f_\ast \mid \bx, \by, \bx_\ast) =
\int p(f_\ast \mid \bbf) p( \bbf \mid \by) \dd \bbf
\label{eq:postpred}
\end{align}
(see \citet{bishop2007pattern}) 
where the $\mathcal{GP}(m,k)$ prior gives the conditional distribution
\begin{align}
\label{eq:postprior}
p(f_\ast \mid \bbf) = \mathcal{N}(f_\ast \mid m(x) + \bk^T \bK^{-1} ( \bbf - \bbm ), k - \bk^T \bK^{-1} \bk) \, .
\end{align}
We can obtain predictions in two different ways: Either 
by substituting  $q(\bbf)$ for $p( \bbf \mid \by)$ in \eqref{eq:postprior} (this leads to \eqref{eq:gppredmean} and \eqref{eq:gppredbSigma}), or simply as the predictions of the mean
\begin{align*}
    \sigma(\E_q[\bbf \mid \by]) = \sigma(\hat{\bbf}).
\end{align*}
We note that while the two are different in general, the decision boundary is identical for binary classification \citep[Sec. 10.3]{bishop2007pattern}.

\subsection{Data-centric self-distillation for GP classification}
For a single step of data-centric distillation, we can choose either of the above predictions as our new targets and will denote them as $\by_{1}$ irrespective of the choice. However, while $\by \in \{0,1\}^N$ we now have $\by_{1} \in [0,1]^N$, and we can no longer assume a conditional Bernoulli distribution over $\by_{1}$. However, by extending to the continuous Bernoulli distribution \citep{Loaiza-Ganem2019TheAutoencoders}, we can avoid this problem, by introducing the appropriate normalizing constant
\begin{align}\label{eq:c_constant}
    C(\lambda) \eqdef \begin{cases} 2 & \text{if } \lambda = \frac{1}{2}, \\ \frac{2\mathrm{tanh}^{-1}(1-2\lambda)}{1-2\lambda} & \text{otherwise}. \end{cases}
\end{align}

See Figure \ref{fig:cont_bernoulli_constant} for plots of the normalizing constant, its gradient, and Hessian.
We now assume $f_{1} \sim \mathcal{GP}(0,k)$ and 
\begin{align*}
    \by_{1} \mid \bbf_{1} \sim \prod_{n=1}^N C(\sigma(f_{1,n}))\sigma(f_{1,n})^{y_{1,n}}(1-\sigma(f_{1,n}))^{1-y_{1,n}},
\end{align*}
which in turn yields that
\begin{align*}
    \log p(\by_{1} \mid \bbf_{1}) = \by_{1}^\trans \bbf_{1} - \sum_{n=1}^N \log(1+\exp(f_{1,n})) + \sum_{n=1}^N \log C(\sigma(f_{1,n})),
\end{align*}
where we note that $\sum_{n=1}^N \log C(\sigma(f_{1,n})) > 0$. In Proposition \ref{prop:C_derivatives} we find the derivative and second derivative of $\log C(\sigma(a))$.
Thus, if we denote $\bC_{1} \eqdef \sum_{n=1}^N \log(C(\sigma(f_{1,n})))$, we get that the gradient of the conditional density is
\begin{align*}
    \nabla \psi_{1}( \bbf_{1}) &= \by_{1} - \bsigma_{1} - \bK^{-1} \bbf_{1}  + \nabla\bC_{1},
\end{align*}
where we can compute the last term by use of the derivative above.
Similarly, it follows that the Hessian can be computed as
\begin{align*}
    -\nabla \nabla \psi_{1}( \bbf_{1}) &= \bW_{1} + \bK^{-1} - \nabla\nabla\bC_{1},
\end{align*}
where we note that $\nabla\nabla\bC_{1}$ is a diagonal matrix, and that $\bsigma_{1}$ and $\bW_{1}$ are defined analogously to the classical case. With the gradient and Hessian, we can follow the usual setup and compute the MAP estimate for the mean of the Laplace approximation, and use the inverse Hessian as covariance. That is, we use
\begin{align*}
    \bbf_{1}^{\text{new}} &= \bbf_{1} - \left(-\bW_{1} - \bK^{-1} + \nabla\nabla\bC_{1} \right)^{-1} \left(\by_{1} - \bsigma_{1} - \bK^{-1} \bbf_{1}  + \nabla\bC_{1}\right) \\
    &= \left(\bW_{1} + \bK^{-1} - \nabla\nabla\bC_{1} \right)^{-1} \left(\left(\bW_{1} + \bK^{-1} - \nabla\nabla\bC_{1} \right)\bbf_{1} +\by_{1} - \bsigma_{1} - \bK^{-1} \bbf_{1}  + \nabla\bC_{1}\right) \\
    &= \bK\bK^{-1}\left(\bW_{1} + \bK^{-1} - \nabla\nabla\bC_{1} \right)^{-1} \left(\left(\bW_{1} - \nabla\nabla\bC_{1} \right)\bbf_{1} +\by_{1} - \bsigma_{1}  + \nabla\bC_{1}\right) \\
    &= \bK\left(\left(\bW_{1} - \nabla\nabla\bC_{1}\right)\bK + \bI\right)^{-1} \left(\left(\bW_{1} - \nabla\nabla\bC_{1}\right)\bbf_{1} + \by_{1} - \bsigma_{1} + \nabla\bC_{1}\right)
\end{align*}
That is, we now have the Laplace approximation
\begin{align*}
    p(\bbf_{1} \mid \by) \approx q(\bbf_{1} \mid \by) = \mathcal{N}\left(\bbf_{1} \mid \hat{\bbf}_{1}, \left(\bW_{1} + \bK^{-1} - \nabla\nabla\bC_{1} \right)^{-1} \right),
\end{align*}
where $\hat{\bbf}_{1}$ is the MAP estimate from above, and we can also use the Laplace approximation to get the marginal log-likelihood. First by a Taylor approximation of $\psi_{1}(\bbf_{1})$ at $\hat{\bbf}_{1}$, we get that $\psi_{1}(\bbf_{1}) \approx \psi_{1}(\hat{\bbf}_{1}) - \frac{1}{2}(\bbf_{1} - \hat{\bbf}_{1})^\trans \bH^{-1}(\bbf_{1} - \hat{\bbf}_{1})$, where $\bH^{-1}$ is the covariance matrix of $q$, i.e. $\bH \eqdef \bW_{1} + \bK^{-1} - \nabla\nabla\bC_{1}$. Thus, we have that
\begin{align*}
    \log p(\by_{1}) &= \int \psi(\bbf_{1}) d\bbf_{1} \\
    &\approx \psi_{1}(\hat{\bbf}_{1}) + \int - \frac{1}{2}(\bbf_{1} - \hat{\bbf}_{1})^\trans \bH^{-1}(\bbf_{1} - \hat{\bbf}_{1}) d\bbf_{1} \\
    &= \psi_{1}(\hat{\bbf}_{1}) + \frac{N}{2}\log(2\pi) - \frac{1}{2}\log|\bH| \\
    &= \by_{1}^\trans \hat{\bbf}_{1} - \sum_{n=1}^N \log(1+\exp(\hat{f}_{1,n})) + \sum_{n=1}^N \log C(\sigma(\hat{f}_{1,n})) - \frac{1}{2}\hat{\bbf}_{1}^\trans \bK^{-1}\hat{\bbf}_{1} \\
    &\qquad- \frac{1}{2}\log|\bK| - \frac{N}{2}\log(2\pi)  + \frac{N}{2}\log(2\pi) - \frac{1}{2}\log|\bH| \\
    &= \by_{1}^\trans \hat{\bbf}_{1} - \sum_{n=1}^N \log(1+\exp(\hat{f}_{1,n})) + \sum_{n=1}^N \log C(\sigma(\hat{f}_{1,n})) \\
    &\qquad - \frac{1}{2}\hat{\bbf}_{1}^\trans \bK^{-1}\hat{\bbf}_{1} - \frac{1}{2}\log(|\bK|) - \frac{1}{2}\log(|\bW_{1} + \bK^{-1} - \nabla\nabla\bC_{1}|)
\end{align*}
The marginal log-likelihood is useful for hyperparameter optimization.

We can then iterate this distillation procedure any number of times. Usually, we consider the inversion of $\bK$ the most computationally expensive step of fitting a GP, and since we can re-use $\bK^{-1}$ for all our steps, the addition of distillation steps is, computationally, not very demanding. In the above MAP estimation, we even rewrite the optimization to avoid the use of the inverse of $\bK$.

\section{Setup for toy-examples and additional results}\label{sec:example_details}
Throughout the main paper we have included illustrations and plots based on various toy-examples. However, all these illustratory examples have been computed with a radial basis kernel function
\begin{align*}
    k(\bx_1, \bx_2) = \sigma_f^2 \exp\left(-\frac{\norm{\bx_1 - \bx_2}^2}{2l} \right),
\end{align*}
where $\sigma_f^2 > 0$ and $l > 0$ are both hyperparameters tunable for the particular example. We also (mainly for the regression setup) consider a noise parameter $\gamma \geq 0$ which is added to the diagonal of $\bK$, i.e. assume noise on the training observations. Even without this assumptions we typically use a small $\gamma \approx 10^{-8}$ to avoid numerical instability both in regression and classification examples.

\begin{figure}[htbp]
    \centering
    \begin{subfigure}[b]{0.48\textwidth}
        \centering
        \includegraphics[width=\linewidth]{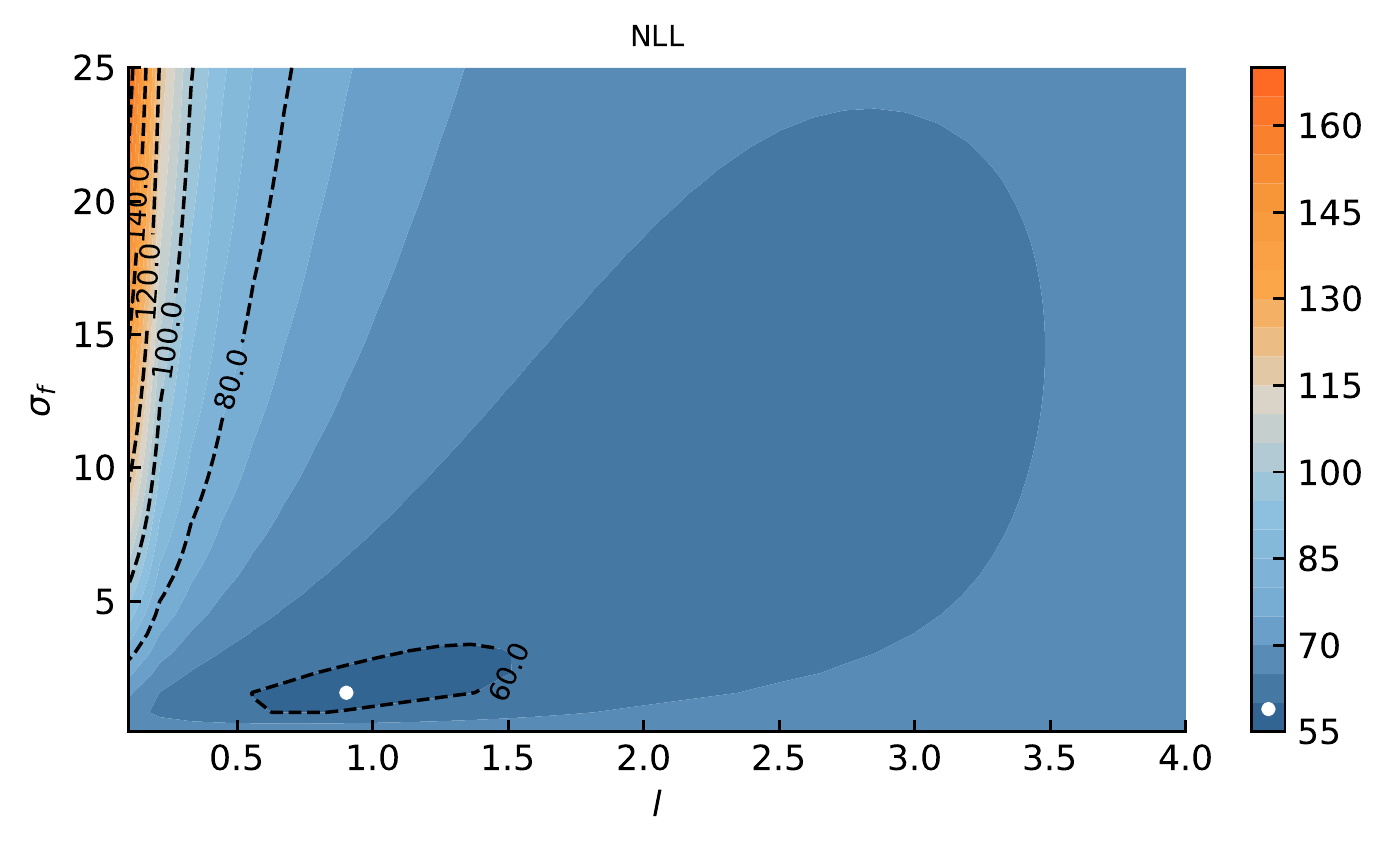}
        \caption{Ordinary GPC}
    \end{subfigure}
    \hfill
    \begin{subfigure}[b]{0.48\textwidth}
        \centering
        \includegraphics[width=\linewidth]{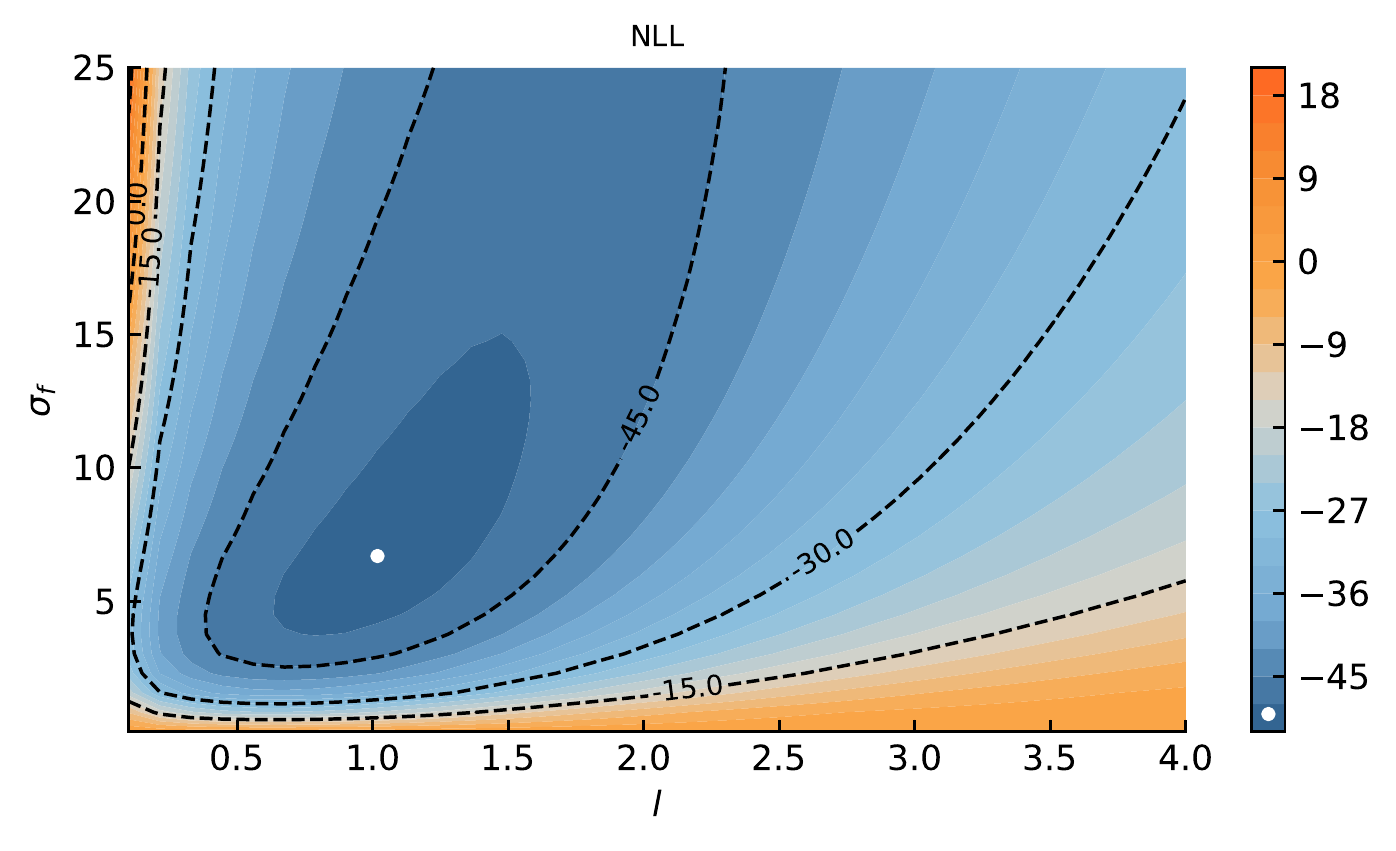}
        \caption{C-GPC}
    \end{subfigure}
    \caption{Grid search of hyperparameters of two different types of GP classification. We indicate the minimum of each grid with a white dot. Note, the absolute values can not be compared between the two models due to different likelihood functions.}
    \label{fig:grid_search_gpc}
\end{figure}

In Figure \ref{fig:grid_search_gpc} we plot the negative log-likelihood over a grid of values for $\sigma_f$ and $l$ both in the case of using the ordinary GP classification with Bernoulli likelihood and the adjusted setting with continuous Bernoulli likelihood, when the input are the true continuous underlying function $g(x) = 2\sin(x\pi/2)$.

\subsection{Additional experiments and results}
\begin{figure}[htbp]
    \centering
    \begin{subfigure}[b]{0.48\textwidth}
        \centering
        \includegraphics[width=\linewidth]{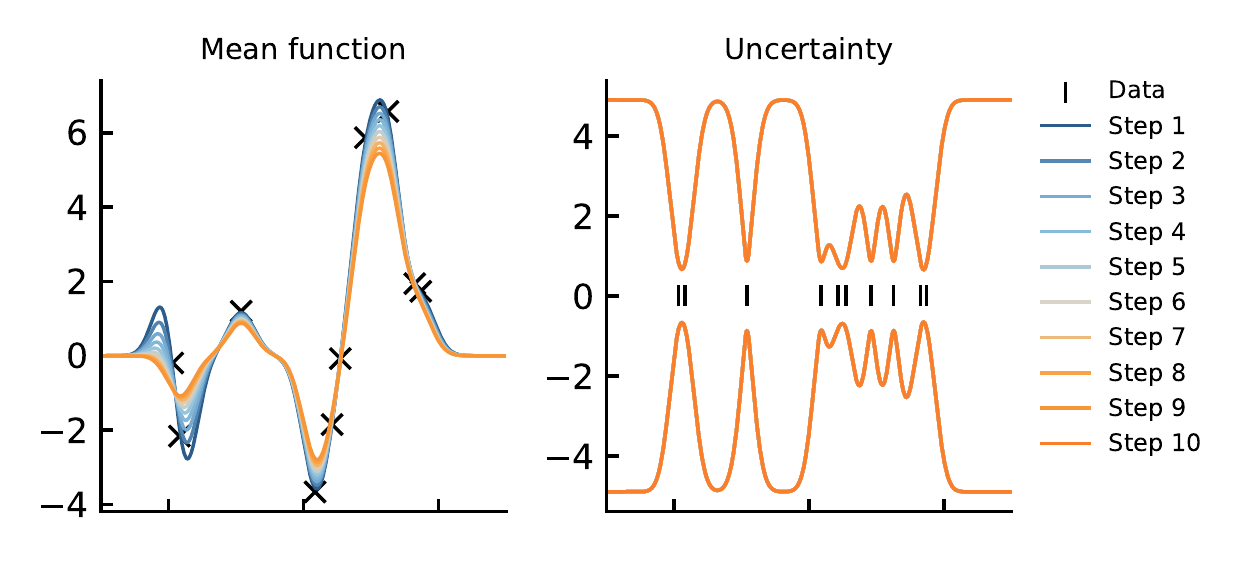}
        \caption{Noise parameters: $(\gamma_1, \dots, \gamma_{10}) = (0.2, \dots, 0.2)$}
    \end{subfigure}
    \hfill
    \begin{subfigure}[b]{0.48\textwidth}
        \centering
        \includegraphics[width=\linewidth]{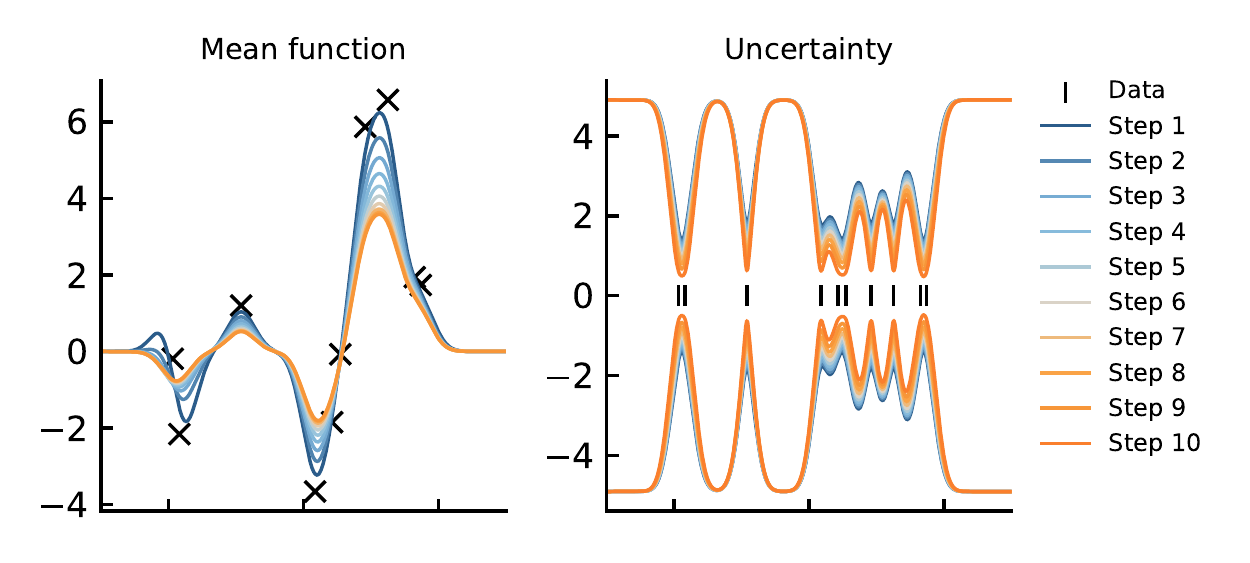}
        \caption{Noise parameters: $(\gamma_1, \dots, \gamma_{10}) = (1, \dots, 0.1)$}
    \end{subfigure}
    \begin{subfigure}[b]{0.48\textwidth}
        \centering
        \includegraphics[width=\linewidth]{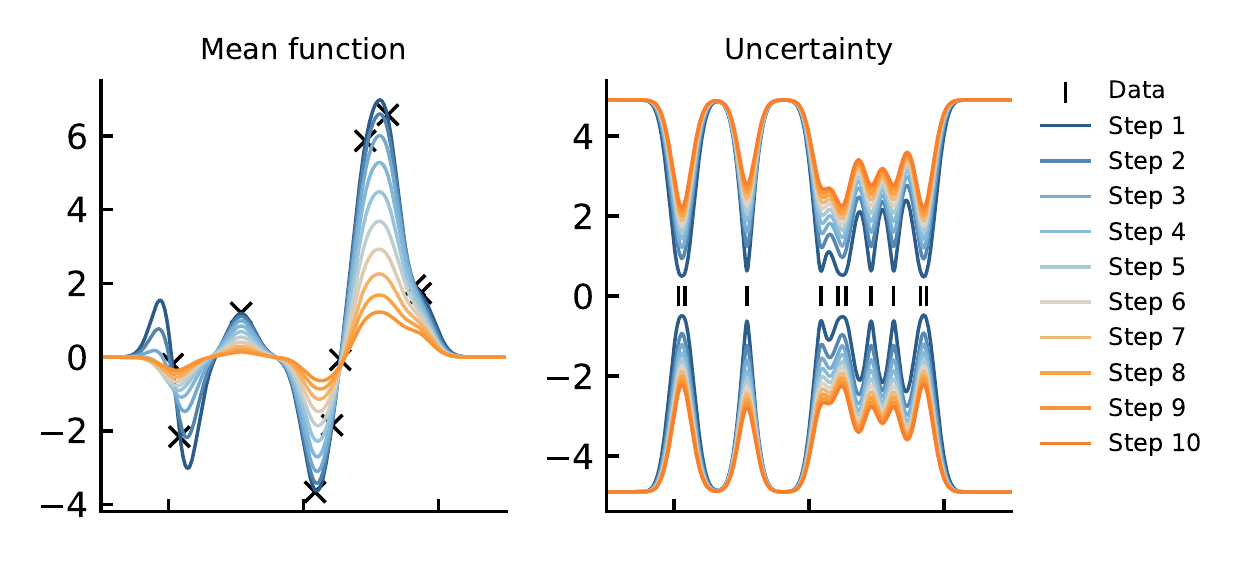}
        \caption{Noise parameters: $(\gamma_1, \dots, \gamma_{10}) = (0.1, \dots, 3)$}
    \end{subfigure}
    \hfill
    \begin{subfigure}[b]{0.48\textwidth}
        \centering
        \includegraphics[width=\linewidth]{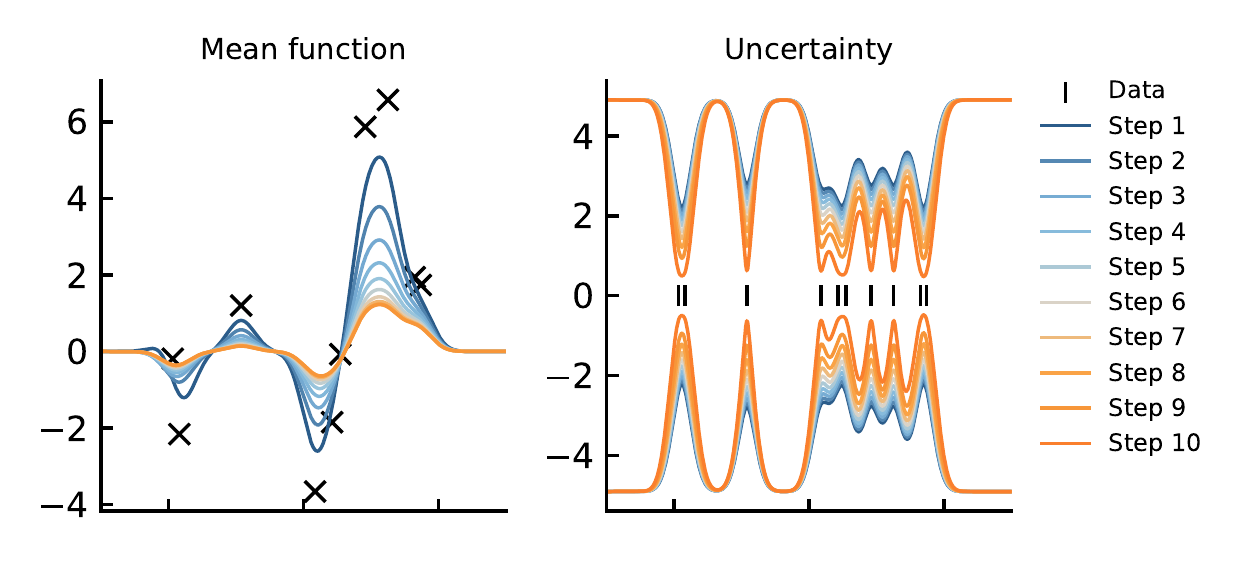}
        \caption{Noise parameters: $(\gamma_1, \dots, \gamma_{10}) = (3, \dots, 0.1)$}
    \end{subfigure}
    \caption{Ten steps of data-centric self-distillation for GPR with fixed hyperparameters, but with varying choices of noise parameters. We note the progressively increasing regularization present in all examples.}
    \label{fig:extra_example_data_gpr}
\end{figure}

\begin{figure}[htbp]
    \centering
    \begin{subfigure}[b]{0.48\textwidth}
        \centering
        \includegraphics[width=\linewidth]{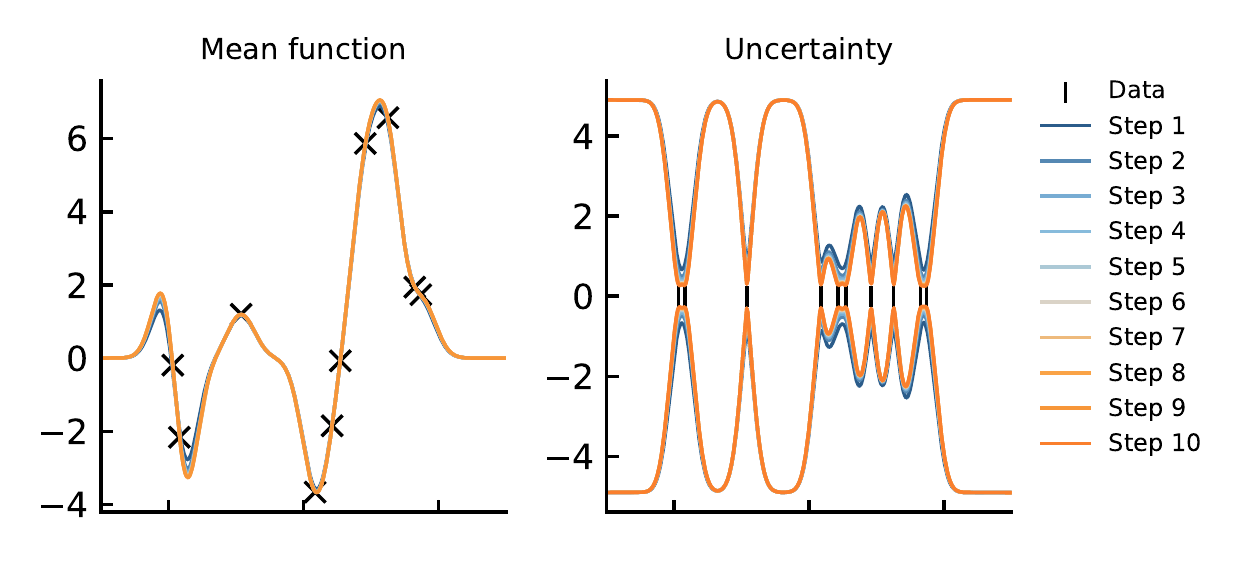}
        \caption{Noise parameters: $(\gamma_1, \dots, \gamma_{10}) = (0.2, \dots, 0.2)$}
    \end{subfigure}
    \hfill
    \begin{subfigure}[b]{0.48\textwidth}
        \centering
        \includegraphics[width=\linewidth]{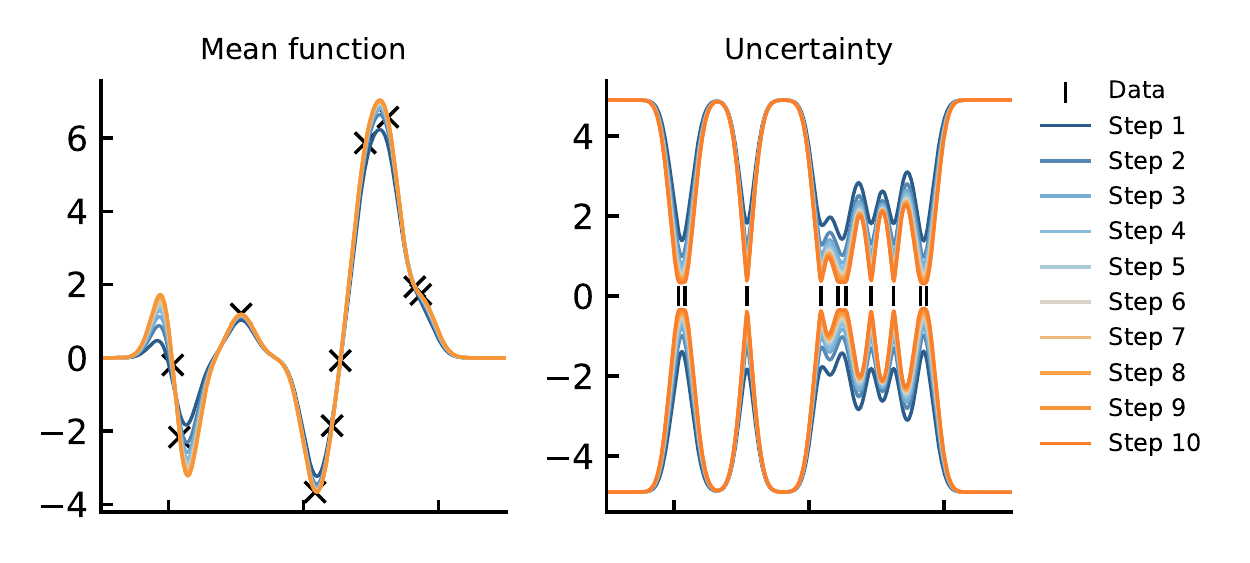}
        \caption{Noise parameters: $(\gamma_1, \dots, \gamma_{10}) = (1, \dots, 0.1)$}
    \end{subfigure}
    \begin{subfigure}[b]{0.48\textwidth}
        \centering
        \includegraphics[width=\linewidth]{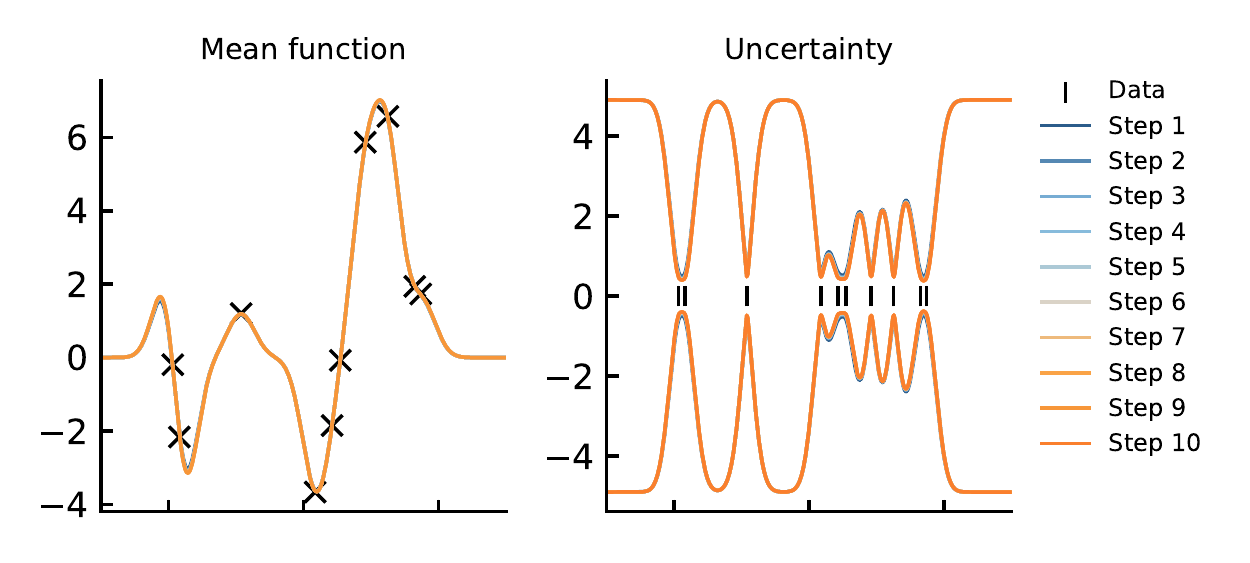}
        \caption{Noise parameters: $(\gamma_1, \dots, \gamma_{10}) = (0.1, \dots, 3)$}
    \end{subfigure}
    \hfill
    \begin{subfigure}[b]{0.48\textwidth}
        \centering
        \includegraphics[width=\linewidth]{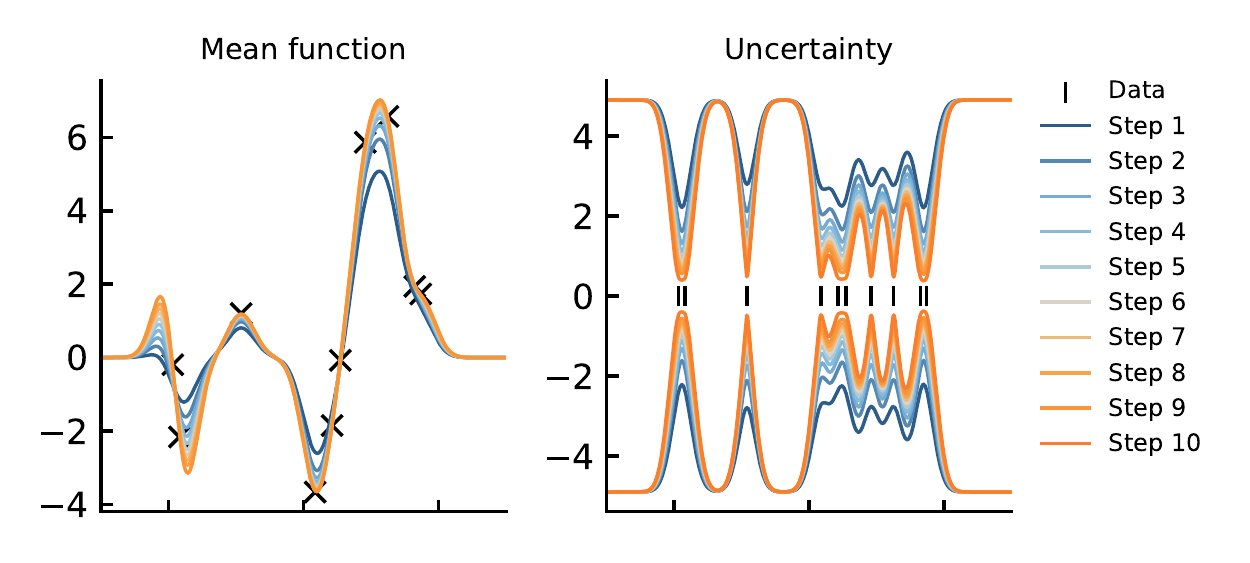}
        \caption{Noise parameters: $(\gamma_1, \dots, \gamma_{10}) = (3, \dots, 0.1)$}
    \end{subfigure}
    \caption{Ten steps of distribution-centric self-distillation for GPR with fixed hyperparameters, but with varying choices of noise parameters. We note that while (c) and (d) yield the same solution at step 10, the path to that solution varies significantly.}
    \label{fig:extra_example_distribution_gpr}
\end{figure}

\begin{figure}
    \centering
    \includegraphics[width=0.5\linewidth]{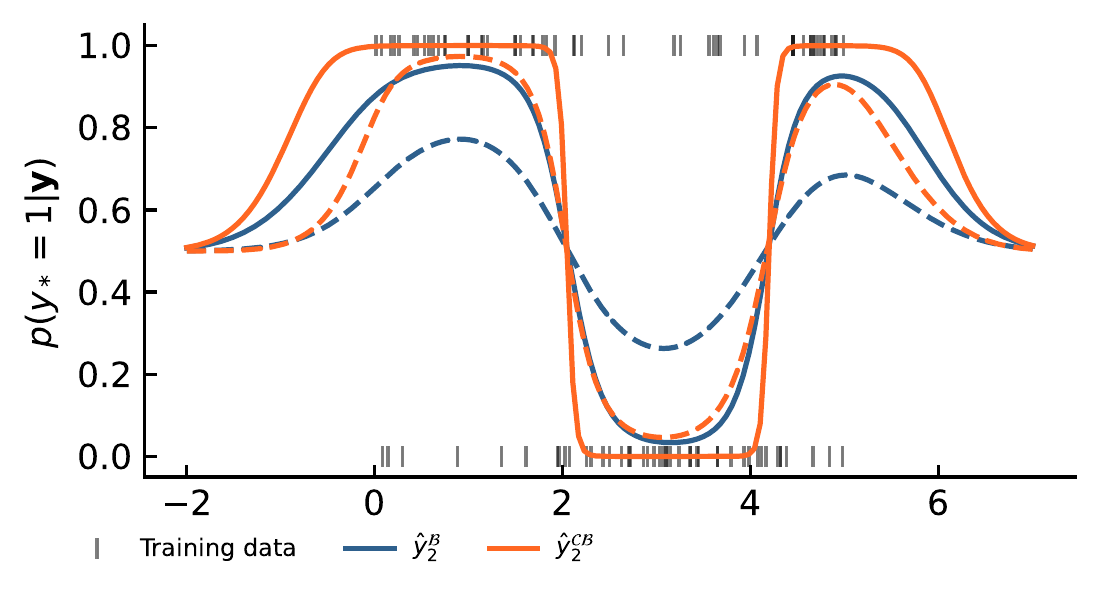}
    \caption{Same setup as in Figure \ref{fig:GPC_data_example_cb_vs_b}, but instead of fitting the distilled model to the soft output of the first, we also include (in dashed) the ordinary GPC and C-GPC when fitted to the binary predictions (thresholded at 0.5). We see that the overfitting of the C-GPC is amplified, while the ordinary GPC closely resembles the initial C-GPC on soft targets.}
    \label{fig:extra_examples_hard_label_data_gpc}
\end{figure}

In the following, we present additional results on the same illustratory experiments as in the main paper, but with changes to e.g. the model parameters or data-assumptions. In particular, in Figure \ref{fig:extra_example_data_gpr} we present different solutions to the data-centric GPR example in the main text, when varying the choices of $(\gamma_1, \dots, \gamma_{10})$. Similarly, in Figure \ref{fig:extra_example_distribution_gpr} we repeat the same experiments for distribution-centric GPR.
Finally, in Figure \ref{fig:extra_examples_hard_label_data_gpc} we plot the solutions of data-centric GPC, when we fit the distilled model, not to the continuous predictions of the initial model, but to the $(0,1)$-encoded targets obtained by thresholding the initial model at $0.5$. In this setting, using a ordinary Bernoulli assumption does not result in a misspecified model, but the thresholded neglects some of the information contained in the continuous targets, and as evident from Figure\ref{fig:extra_examples_hard_label_data_gpc}, the distilled models in this case is largely similar to that of a C-GPC on the continuous targets.

\section{Some details on the continuous Bernoulli distribution}\label{sec:cont_bern_details}
The continuous Bernoulli distribution (denoted by $\mathcal{CB}(\lambda)$) was introduced in \citet{Loaiza-Ganem2019TheAutoencoders} and dealt with the relatively common case, where data is continuous values in $[0,1]$ rather than discrete values in $\{0,1\}$. When performing distillation this is indeed the case, and merely using a (implicit or explicit) Bernoulli assumption is not suitable. For good measure, we repeat some key results of $\mathcal{CB}$ from \citet{Loaiza-Ganem2019TheAutoencoders} in the following section. Note, in the following, we will treat Bernoulli variables as if they could attain continuous values in $[0,1]$, which is the incorrect approach currently often used in practice.

\begin{figure}[htbp]
    \centering
    \includegraphics[width=0.9\linewidth]{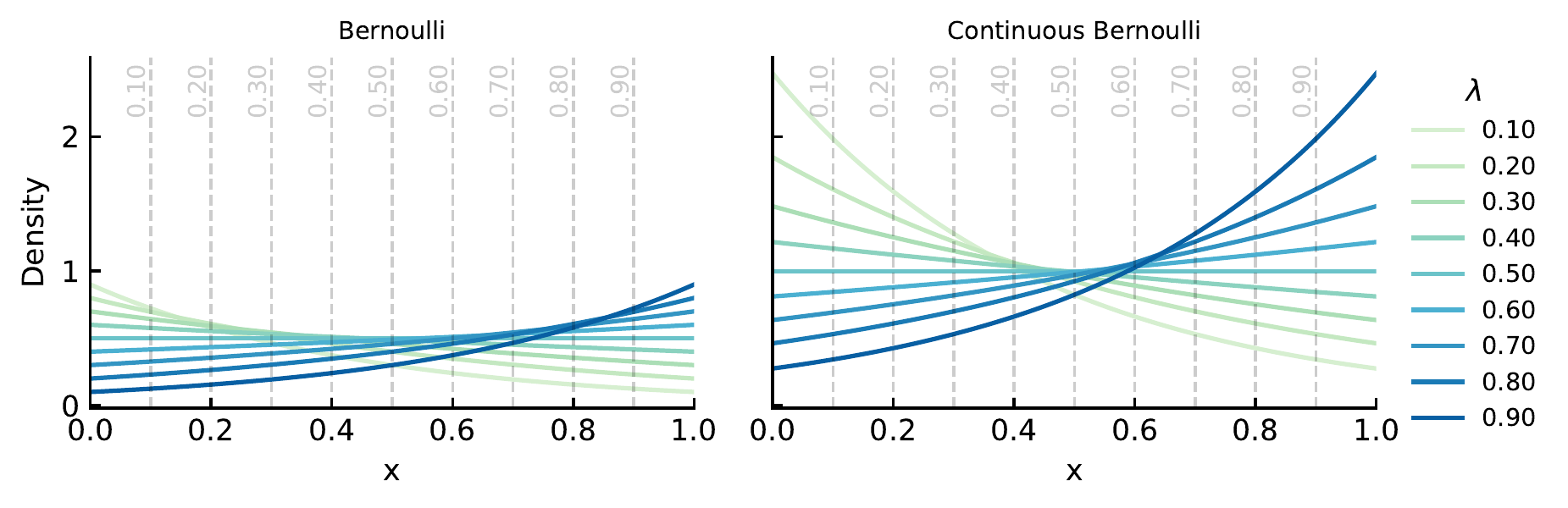}
    \caption{Density of Bernoulli and continuous Bernoulli distributions with different choices of parameter $\lambda$.}
    \label{fig:cont_bernoulli_density}
\end{figure}

First, we recall that the density of the Bernoulli distribution with parameter $\lambda \in (0,1)$ is given by
\begin{align*}
    X \sim \mathcal{B}(\lambda) \iff \tilde{p}(x \mid \lambda) = \lambda^{x}(1-\lambda)^{1-x}.
\end{align*}
Second, by \citet{Loaiza-Ganem2019TheAutoencoders} the density of $\mathcal{CB}$ differs from the density of $\mathcal{B}$ merely by a multiplicative normalizing constant (dependent on the parameter $\lambda$), denoted by $C(\lambda)$. See Figure \ref{fig:cont_bernoulli_density} for a plot of both densities. In particular, we have that for $\lambda \in (0, 1)$ then
\begin{align*}
    X \sim \mathcal{CB}(\lambda) \iff p(x \mid \lambda) = C(\lambda)\lambda^{x}(1-\lambda)^{1-x}
\end{align*}
where the normalizing constant can be shown to be
\begin{align*}
    C(\lambda) \eqdef \begin{cases}
    2 & \text{if } \lambda = \frac{1}{2}, \\
    \frac{2\mathrm{tanh}^{-1}(1-2\lambda)}{1-2\lambda} & \text{otherwise},
    \end{cases}
\end{align*}
and $\mathrm{tanh}^{-1}$ is the inverse hyperbolic tangent function.

\begin{figure}[htbp]
    \centering
    \includegraphics[width=0.9\linewidth]{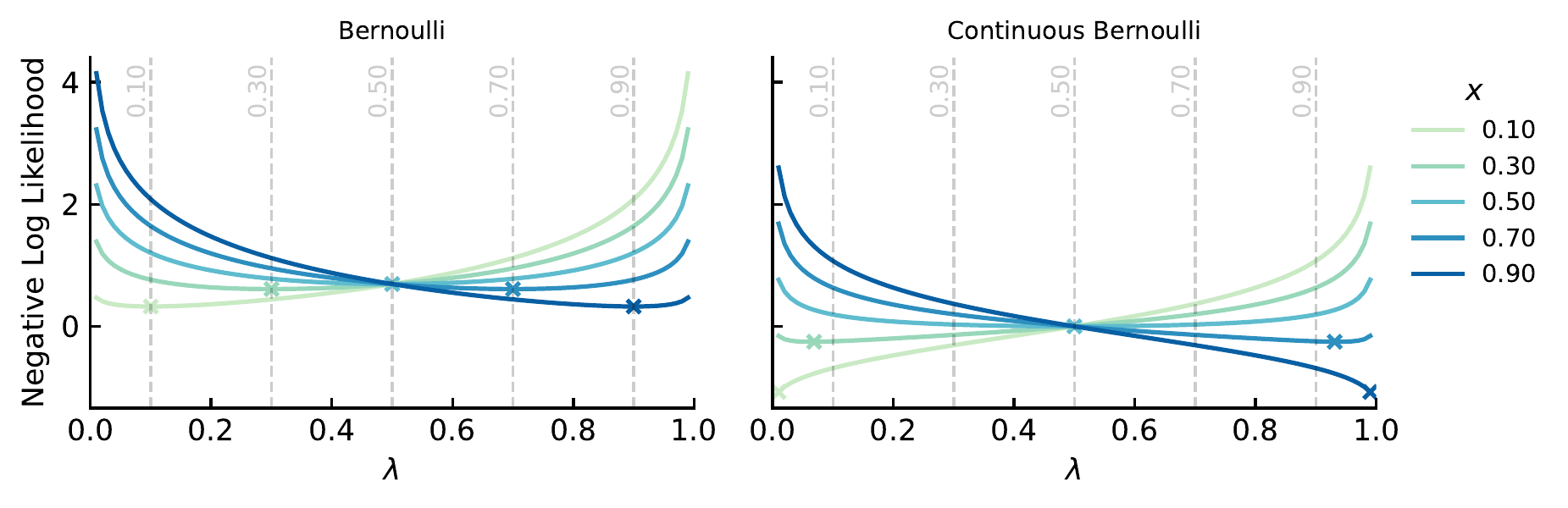}
    \caption{Negative log-likelihood of Bernoulli and continuous Bernoulli distributions with different choices of parameter $\lambda$. We illustrate the optimizing $\lambda$ for each $x$ by a cross $\times$, and note that for a given observation $x$ the optimal choice of $\lambda$ for the Bernoulli is $x$, while it for the continuous Bernoulli is a non-linear function of $x$, which is generally biased towards more extreme values of $\lambda$. See also Figure \ref{fig:cont_bernoulli_probs}.}
    \label{fig:cont_bernoulli_likelihood}
\end{figure}

\begin{figure}[htbp]
    \centering
    \includegraphics[width=0.8\linewidth]{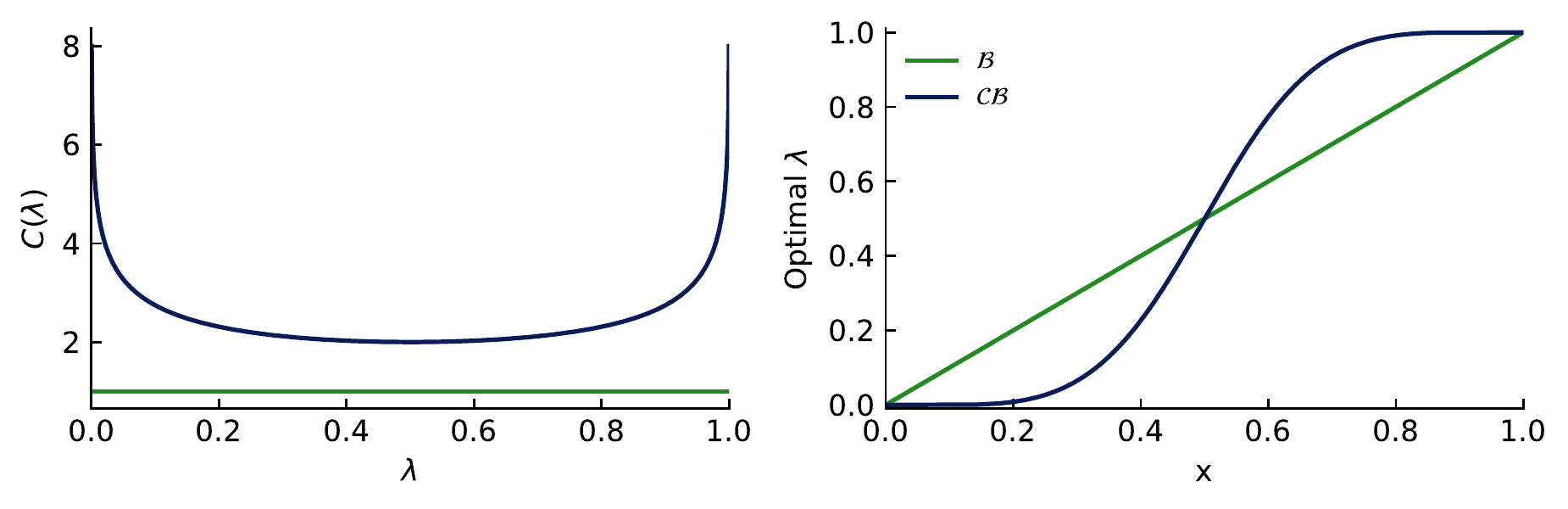}
    \caption{Normalizing constant and optimal $\lambda$ values (i.e. maximizing likelihood for a single observation of $x$) for both ordinary and continuous Bernoulli distributions.}
    \label{fig:cont_bernoulli_probs}
\end{figure}

In Figure \ref{fig:cont_bernoulli_probs} we plot the normalizing constant for both the ordinary (equal to $1$ for all $\lambda$) and continuous Bernoulli distribution, and we also observe that since the mean of a $\mathcal{CB}(\lambda)$ distributed variable is not merely $\lambda$, the optimal $\lambda$ (i.e. maximizing the density) for a single observation $x$ is different from $\lambda$.

\begin{figure}[htbp]
    \centering
    \includegraphics[width=0.9\linewidth]{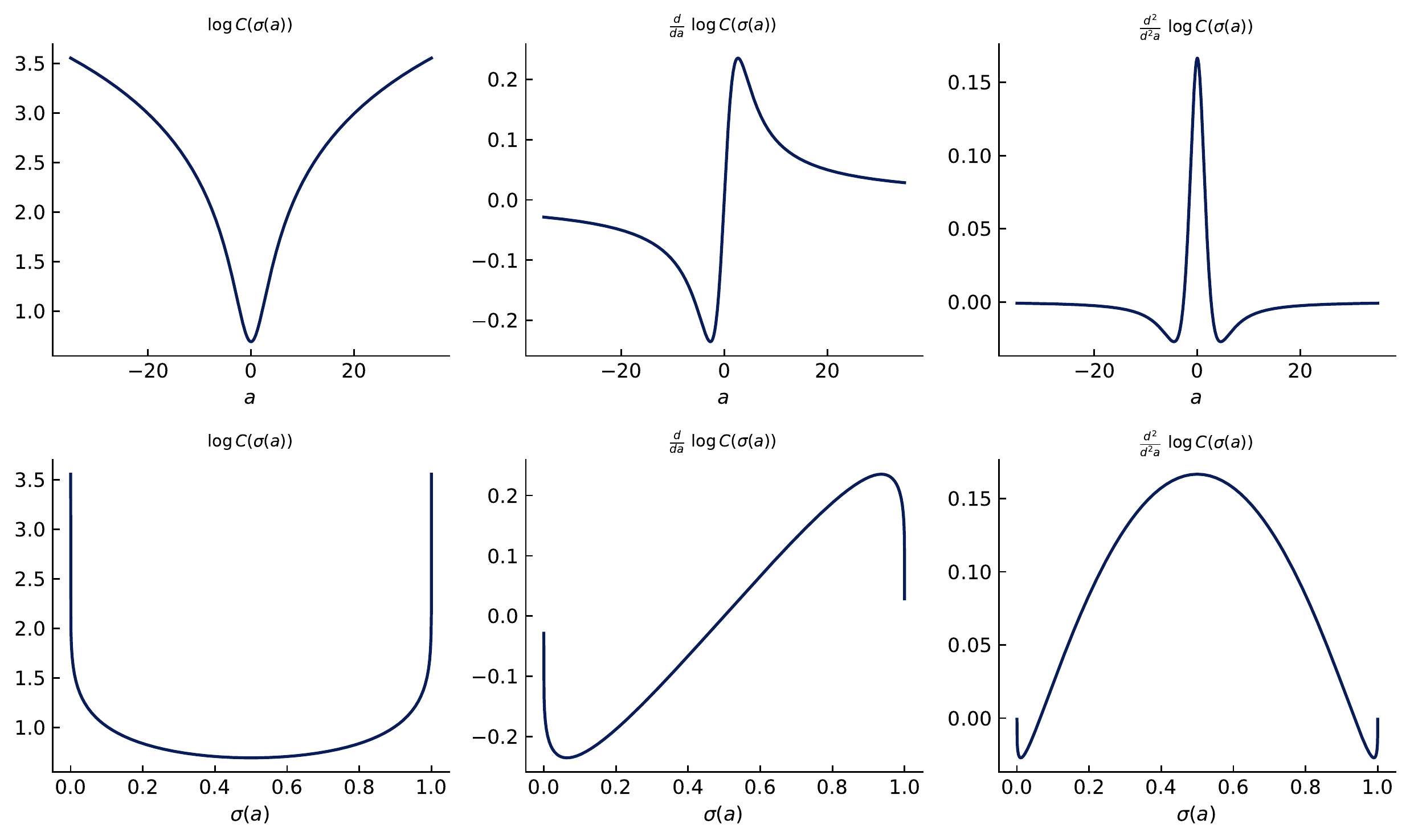}
    \caption{Log-normalizing constant for the continuous Bernoulli distribution $C(\lambda)$, along with first and second derivatives wrt. $a$.}
    \label{fig:cont_bernoulli_constant}
\end{figure}

Finally, in Figure \ref{fig:cont_bernoulli_constant} we both plot the log-normalizing constant, as well as the first and second derivatives wrt. $a$ versus either $\lambda = \sigma(a)$ or $a$. See Proposition \ref{prop:C_derivatives} for the derivations of these.
\section{Time and complexity analysis}
Despite the main aim of our proposed self-distillation methods being of a theoretical nature, in the following, we investigate the compute requirements of our proposed methods.

\subsection{(Efficient) Implementation}
We provide a public implementation of our methods in python on \url{github.com/Kennethborup/gaussian_process_self_distillation}. We provide four classes; data-centric and distribution-centric self-distillation for both GP regression and GP classification. The classes are named straightforwardly as \texttt{DataCentricGPR}, \texttt{DistributionCentricGPR}, \texttt{DataCentricGPC}, and \texttt{DistributionCentricGPC}. Each method has \texttt{.fit}, and \texttt{.predict} methods that are similar to the scikit-learn API. For \texttt{DataCentricGPR} we provide both a naïve and efficient implementation, which significantly improves the training speed for multiple steps of distillation. In particular, inspired by \citet{Borup2021} we utilize the singular values decomposition of $\bK$ as $\bV\bD\bV^\trans$ (here $\bV \in \R^{N \times N}$ is an orthogonal matrix with the eigenvectors of $\bK$ as rows and $\bD \in \R^{N \times N}$ is a non-negative
diagonal matrix with the associated eigenvalues in the diagonal), to rewrite \eqref{eq:kernel_sd} for $t=1$ into
\begin{align*}
    \by_1 & = \bV \bD\left(\bD + \gamma_1\bI_N \right)^{-1}\bV^\trans \by
\end{align*}
which by iterative inserting yields the more general expression
\begin{align*}
    \by_t &= \left(\prod_{s=1}^{t}\bK(\bK + \gamma_s \bI_N)^{-1}\right)\by, \\
    &= \bV\left(\prod_{s=1}^{t}\bA_{s}\right) \bV^\trans\by, 
\end{align*}
where $\bA_s = \bD (\bD + \gamma_s \bI_N)^{-1} = \diag_{n=1, \dots,. N}\left(\frac{d_n}{d_n+\gamma_s} \right)$ for $s = 1, \dots t$ are diagonal matrices. Thus, when $\bV$ and $\bD$ are computed, the subsequent distillation steps are computationally cheap due to the diagonal structure of the $\bA_s$'s. One can similarly rewrite the prediction formula \eqref{eq:kernel_sd_pred} - see the source code for more implementation details.

\subsection{Time requirements}
\begin{figure}[htbp]
    \centering
    \includegraphics[width=0.9\linewidth]{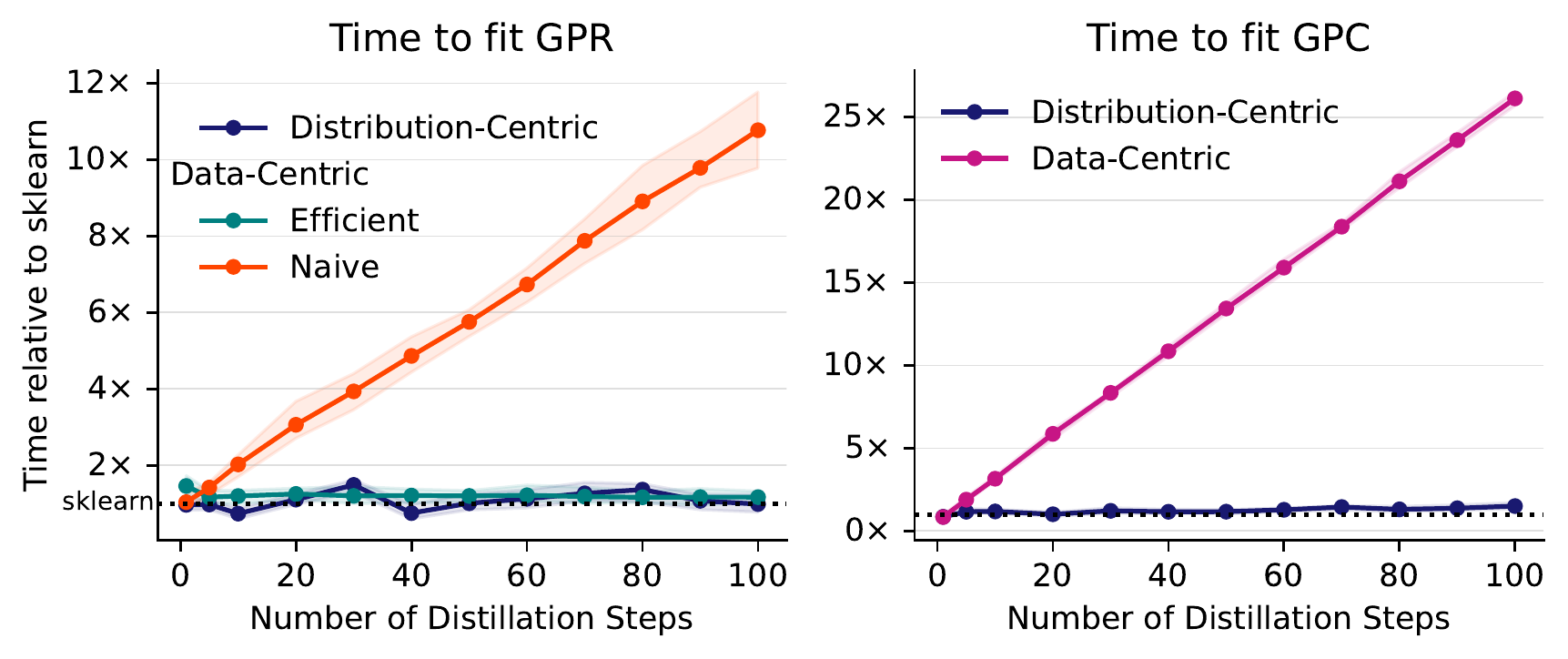}
    \caption{Relative time to fit various self-distilled GP models compared to a single fit with the standard \texttt{sklearn} implementation of the regression and classification model respectively. On the (left) we plot the relative time to fit regression models. For data-centric self-distillation, we use both a naïve and efficient approach. While the naïve approach scales linearly with the distillation steps, the efficient approach is constant in relative time for all distillation steps. On the (right) we plot the relative time for the classification models compared to the \texttt{sklearn} implementation and observe that the data-centric approach scales linearly with distillation steps, while (as expected) the distribution-centric approach is constant with distillation steps. The shaded regions represent the 10 and 90 empirical quantiles of 30 repetitions for each method and the number of distillation steps.}
    \label{fig:time_to_fit_relative}
\end{figure}

To evaluate the practical implications of our proposed methods, we compared the time it takes to fit our methods (for a varying number of distillation steps) with an ordinary fit using the standard implementation of GP regression and GP classification in \texttt{sklearn}; namely the \texttt{GaussianProcessRegressor} and \texttt{GaussianProcessClassifier}. In particular, we fit a model with each of our methods for a particular number of steps and divide the fitting time with the fitting time of the corresponding ordinary \texttt{sklearn} method. We repeat our experiments 30 times and report the mean and 10 and 90 empirical quantiles in Figure \ref{fig:time_to_fit_relative}. All experiments are performed on a Mac M1 Pro CPU.

For regression, we observe that distribution-centric self-distillation (as expected from Theorem \ref{th:mainrec}) requires no more computation than ordinary GPR with the \texttt{sklearn} implementation of GP regression. This is simply due to distribution-centric self-distillation for regression stays within the GP setting. For data-centric self-distillation the story is different; the time complexity of a naïve iterative implementation where each successive model is fitted to the predictions of the preceding model scales linearly with the number of distillation steps (at a rate of $\approx 0.11$). However, by utilization of an eigendecomposition of $\bK$ in \eqref{eq:kernel_sd}, one can rewrite the solution to optimize the fitting speed significantly. In particular, each additional step of distillation merely requires updating a diagonal matrix and a few matrix products (of re-used matrices), which is much cheaper. Thus, with this efficient implementation, we are able to fit data-centric self-distillation to any number of distillation steps at a constant time complexity. Thus, both data-centric and distribution-centric self-distillation for regression does not take longer to fit than an ordinary GP regression from \texttt{sklearn}.

For classification, we observe that data-centric self-distillation scales linearly with the number of distillation steps at a rate of $\approx 0.26$, and multiple steps of self-distillation in this setting are quickly computationally costly. However, the time to fit with distribution-centric self-distillation is (as expected) constant with the number of distillation steps. This clearly expectable from Proposition \ref{prop:gpc_scaling}, and fits as quickly as the \texttt{sklearn} implementation.

\subsection{Computational requirements}
In the following, we consider the computational requirements of our methods. We will consider the inversion of matrices, matrix products, Cholesky decomposition, and SVD to all be $\mathcal{O}(N^3)$ ignoring constants, although some algorithms exist to reduce the order of computation for different methods to $\mathcal{O}(N^\omega)$ where typically $2 < \omega \leq 3$. Our implementation is based on Algorithm 3.1, 3.2, and  5.1 in \citet{Rasmussen2006}, which largely depends on using the Cholesky decomposition for back substitution.

We note that while the framing of our setup differs notably from the ordinary GP regression and GP classification setups, our proposed methods still stay close to or within the ordinary GP framework simplifying the analysis. In particular, for distribution-centric GP regression, the solution is an ordinary GPR model with a particularly scaled noise parameter (see Theorem \ref{th:mainrec}), and any existing GPR implementation can be used. Thus, the complexity of this method is $\mathcal{O}(N^3)$ as usual. For distribution-centric GP classification, we consider the approximate method from Proposition \ref{prop:gpc_scaling}, and note, that this too is equivalent to an ordinary GPC model, and is in $\mathcal{O}(N^3)$. For data-centric GP regression, a naive implementation would require $t$ iterative fits of an ordinary GPR model, but utilizing the efficient implementation above, we replace the Cholesky decomposition with the SVD, and can thus reuse $\bV$, and $\bD$ for multiple steps of distillation. Thus, each additional distillation step merely requires matrix products of diagonal matrices and with the $N \times N$ matrices, which makes this method $\mathcal{O}(N^3)$ as well. Finally, due to the iterative nature of data-centric GP classification, which does not allow a simplified solution, this method, unfortunately, scales linearly with the number of distillation steps (although still in $\mathcal{O}(N^3)$). In particular, for $t>1$ we change the likelihood function to the continuous Bernoulli, which is a negligent change due to the simple and closed-form solutions provided in Proposition \ref{prop:C_derivatives}, but since we do not have a closed form solution for our model, we need to fit each model in an ordinary fashion, thus requiring $\mathcal{O}^(tN^3)$ with $t$ the number of distillation steps. This can be computationally demanding for large $t$. Generally, we observe that, despite data-centric GP classification scaling linearly with $t$, all our proposed methods are still primarily limited by the scaling of $N$. Finally, due to the minor changes in our methods, methods to e.g. efficiently compute/approximate $\bK$ should largely still be compatible with our methods.


\end{document}